\documentclass{article}

\PassOptionsToPackage{numbers, sort, compress}{natbib}

\usepackage[preprint]{neurips_2026}

\usepackage[utf8]{inputenc} 
\usepackage[T1]{fontenc}    
\usepackage{hyperref}       
\usepackage{url}            
\usepackage{booktabs}       
\usepackage{amsfonts}       
\usepackage{nicefrac}       
\usepackage{microtype}      
\usepackage{xcolor}         

\usepackage{amsmath}
\usepackage{amssymb}
\usepackage{graphicx}
\usepackage{multirow}
\usepackage{bm}
\usepackage{array}
\usepackage{makecell}
\usepackage{adjustbox}
\usepackage[table]{xcolor}
\usepackage{pifont}
\usepackage{algorithm}
\usepackage{algpseudocode}
\usepackage{wrapfig}
\usepackage{enumitem}

\title{Less is More: Compact-Token Masked Feature Prediction for Skeleton Representation Learning}

\author{%
  Jeonghyeok~Do\\
  Information \& Electronics Research Institute\\
  KAIST\\
  \texttt{ehwjdgur0913@kaist.ac.kr}\\
  \And
  Chen~Yun\\
  School of Electrical Engineering\\
  KAIST\\
  \texttt{cyruby@kaist.ac.kr}\\
  \And
  Geunhyuk~Youk\\
  School of Electrical Engineering\\
  KAIST\\
  \texttt{rmsgurkjg@kaist.ac.kr}\\
  \And
  Munchurl~Kim\thanks{Corresponding author.}\\
  School of Electrical Engineering\\
  KAIST\\
  \texttt{mkimee@kaist.ac.kr} \\
}

\begin{document}

\maketitle

\vspace{-2.0em}
\begin{center}
    \small
    Project page: \url{https://kaist-viclab.github.io/SLiM_site/}
\end{center}
\vspace{0.5em}

\begin{abstract}
Current skeleton representation learning paradigms face distinct limitations: Contrastive Learning (CL) often overlooks fine-grained motion details, while Masked Auto-Encoders (MAE) rely on coordinate-level reconstruction. This reconstruction inherently demands dense token sequences and heavy decoders, wasting pre-training computation on discarded components and forcing downstream inference to process dense token grids. To resolve these bottlenecks, we propose \textbf{SLiM} (\textbf{S}keleton \textbf{L}ess \textbf{i}s \textbf{M}ore), a compact-token framework that unifies masked feature prediction and contrastive learning via a shared encoder. By shifting the objective from raw coordinate reconstruction to decoder-free, teacher-guided feature prediction, SLiM breaks the reliance on dense tokenization and enables effective learning with a highly compact token grid. Crucially, to prevent trivial shortcut learning arising from strong inter-joint dependencies of human, we introduce Semantic Tube Masking together with Skeleton-Aware Augmentations to enforce deep skeletal-temporal reasoning and anatomical consistency. Extensive experiments across multiple downstream protocols demonstrate that SLiM achieves state-of-the-art performance while structurally reducing inference computation by 7.89$\times$ compared to dense-token MAE baselines.
\end{abstract}

\section{Introduction}

Skeleton data provides a compact and informative representation for human action understanding, offering inherent robustness to background and viewpoint changes compared to RGB videos \cite{Review1, Review2, Review3}. While supervised skeleton-based action recognition has achieved strong performance \cite{CTRGCN, Skateformer, PoseC3D}, it depends on costly annotations and often generalizes poorly to unseen domains. Self-supervised learning (SSL) has therefore become an essential direction for learning transferable motion representations from unlabeled sequences \cite{DINO, DINOv2, iBOT, CL, MAE}.

\begin{figure}[t]
  \centering
  \includegraphics[width=0.8\textwidth]{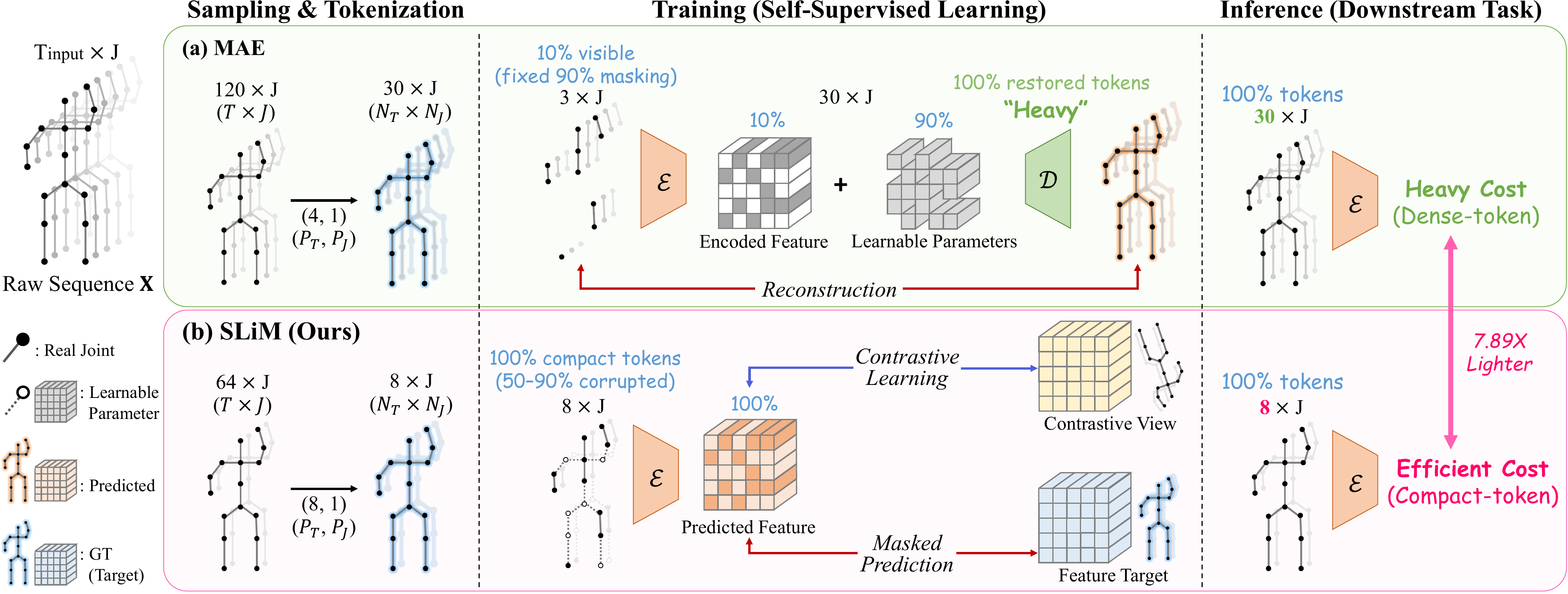}
  \caption{
  Conceptual comparison between coordinate-reconstruction MAE and SLiM.
  (a) Standard skeleton MAE relies on dense skeletal-temporal tokens
  (\(30 \times J\)) and a full-sequence decoder to reconstruct raw coordinates.
  Although the decoder is discarded after pre-training, downstream inference
  still requires dense encoder tokens.
  (b) SLiM replaces coordinate reconstruction with decoder-free masked feature
  prediction under a compact token grid (\(8 \times J\)), using stochastic
  mask-token corruption and contrastive learning during pre-training.
  }
  \label{fig:motiv}
\end{figure}

Existing skeleton SSL methods mainly follow two paradigms: Contrastive Learning (CL) and Masked Auto-Encoding (MAE). CL-based methods \cite{ActCLR, CPM, CMD, AimCLR} learn sequence-level invariance but often overlook fine-grained local patterns, and constructing semantically consistent views is non-trivial for complex actions. MAE-based methods \cite{MAMP, SkeletonMAE, SJEPA, GFP} address this by predicting missing information, but typically rely on fine-grained coordinate reconstruction (Fig.~\ref{fig:motiv}(a)). This inherently demands dense skeletal-temporal tokenization and heavy full-sequence decoders. Furthermore, extreme masking ratios (e.g., 90\%) make the pretext task exceedingly difficult, forcing models to rely on even denser input frames to gather sufficient skeletal-temporal clues. Consequently, this strict dependence on dense tokenization triggers a chain of inefficiencies: wasted pre-training computation on discarded decoders, and a severe inference asymmetry where the encoder must process the entire dense, unmasked token grid.

\begin{wrapfigure}{r}{0.55\columnwidth}
  \centering
  \includegraphics[width=0.50\columnwidth]{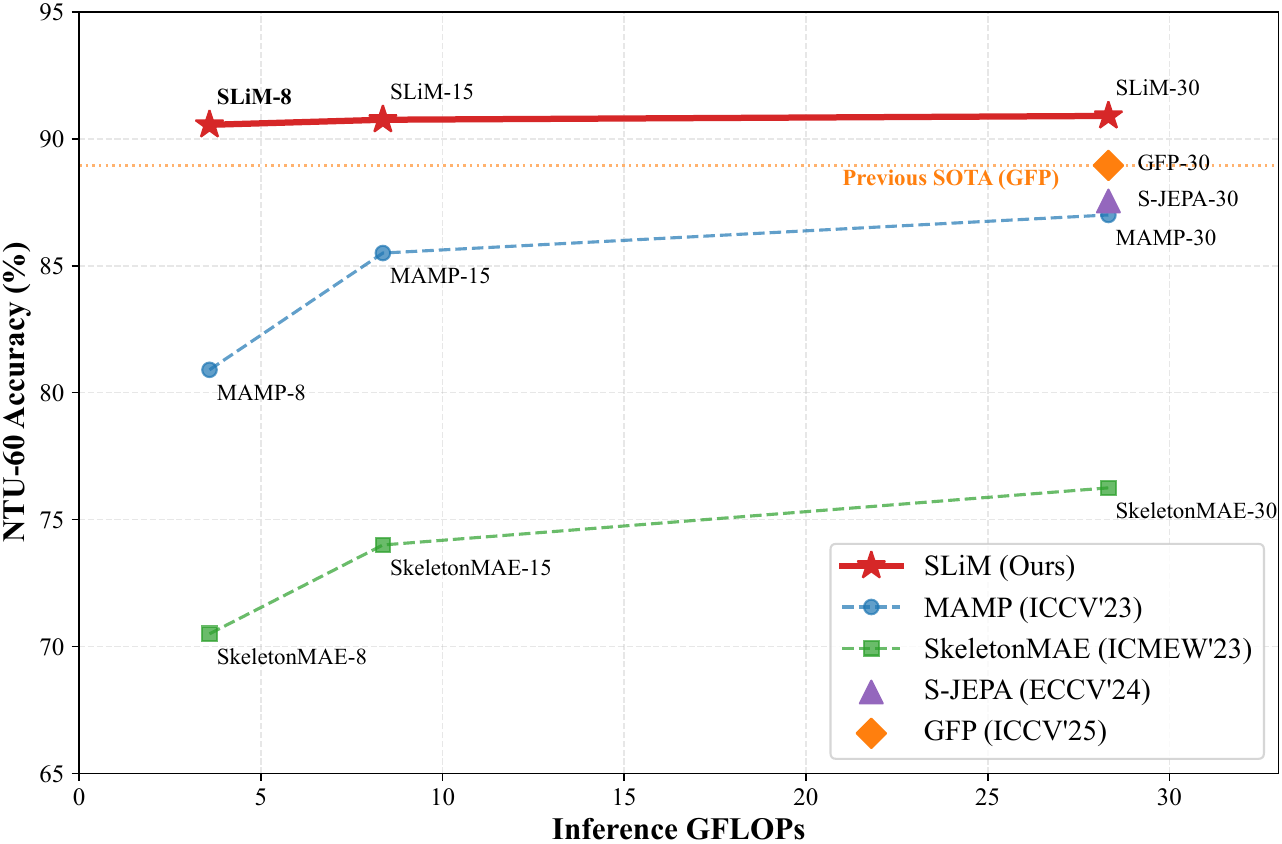}
  \caption{Accuracy--efficiency trade-off on NTU-60. Accuracy is averaged over X-Sub and X-View, and inference cost is measured by downstream GFLOPs. The suffix denotes the temporal token budget ($N_T$). SLiM maintains SOTA accuracy with much lower inference cost than dense-token MAE-based methods.}
  \label{fig:token}
\end{wrapfigure}

As empirically shown in Fig.~\ref{fig:token}, simply reducing the token budget in standard MAEs severely degrades accuracy \cite{MAMP, SkeletonMAE}, confirming their reliance on dense coordinate contexts. To break this bottleneck, we propose SLiM (\textbf{S}keleton \textbf{L}ess \textbf{i}s \textbf{M}ore), a unified SSL framework that combines masked feature prediction and contrastive learning. By reformulating the objective as a decoder-free, teacher-guided feature prediction task (Fig.~\ref{fig:motiv}(b)), SLiM shifts the target from low-level coordinates to high-level semantics. This paradigm shift completely eliminates the need for dense tokenization and heavy decoders. Furthermore, instead of a fixed extreme masking ratio, SLiM applies stochastic mask-token corruption (ranging from 50\% to 90\%) over a compact token grid. This dynamic masking strategy significantly increases the diversity of masked inputs, promoting robust representation learning and explicitly boosting downstream performance. Ultimately, this token-efficient paradigm structurally resolves the computational asymmetry, drastically reducing both pre-training overhead and downstream costs.

However, compact feature prediction alone is vulnerable to shortcut learning; strong skeletal-temporal correlations allow independent joint masking \cite{MAMP, SJEPA, GFP} to be trivially solved via local interpolation. To enforce true action-level reasoning, we introduce \textit{Semantic Tube Masking}, which occludes connected anatomical regions over consecutive frames. For the contrastive branch, we propose \textit{Skeleton-Aware Augmentations} such as skeleton-aware rotation, skeleton-aware mirroring, and bone-aware scaling to generate diverse views while strictly preserving anatomical validity. These components provide challenging masked inputs and semantically reliable contrastive pairs. Our contributions are summarized as follows:

\begin{itemize}
\item We propose \textbf{SLiM}, a compact-token skeleton SSL framework that uses a shared encoder to combine decoder-free masked feature prediction with global–local contrastive learning, completely bypassing the inefficiencies of coordinate reconstruction.
\item We design \textbf{Semantic Tube Masking} and \textbf{Skeleton-Aware Augmentations} to support compact-token learning by avoiding interpolation-based shortcuts and ensuring anatomical consistency.
\item Extensive experiments demonstrate that SLiM achieves SOTA performance across multiple downstream protocols while reducing inference cost by 7.89$\times$ compared to dense-token MAE baselines.
\end{itemize}

\section{Related Work}
\label{sec:related}

\noindent\textbf{Contrastive skeleton SSL.} Contrastive learning (CL) \cite{CL} learns skeleton representations by aligning augmented views of the same sequence. CrosSCLR \cite{CrosSCLR} learns view-invariant features through cross-view consistency, while AimCLR \cite{AimCLR} improves robustness with strong skeleton augmentations. HaLP \cite{HaLP} mines latent positives, and ActCLR \cite{ActCLR} decomposes actions into static and dynamic components. Recent methods such as USDRL \cite{USDRL} incorporate feature decorrelation objectives. Despite their effectiveness, CL-based methods mainly optimize sequence-level alignment, which may miss fine-grained motion patterns and relies on semantically valid views.

\noindent\textbf{Masked and predictive skeleton modeling.} Masked modeling methods learn local skeletal-temporal dependencies by predicting missing skeleton information. SkeletonMAE \cite{SkeletonMAE} reconstructs masked joint coordinates, and MAMP \cite{MAMP} extends this objective to motion-aware trajectory prediction. S-JEPA \cite{SJEPA} moves the prediction target from raw coordinates to latent embeddings, while GFP \cite{GFP} introduces generative feature prediction to connect low-level reconstruction with semantic representation learning. These methods demonstrate the benefit of predictive objectives for skeleton SSL, but they commonly rely on dense skeletal-temporal tokenization, reconstruction or generation modules, and joint/patch-level masking strategies that may exploit strong local correlations among skeleton joints. While decoder-free predictive objectives have been explored in general SSL \cite{DINO, iBOT, DINOv2}, SLiM focuses on their skeleton-specific adaptation under compact token budgets by combining masked feature prediction with structured masking and anatomically consistent augmentations.

\noindent\textbf{Generative and multi-modal pretext tasks.} Beyond standard CL and masked prediction objectives, recent methods have explored generative and multi-modal pretext tasks. MacDiff \cite{MacDiff} formulates skeleton representation learning with masked conditional diffusion, while IGM \cite{IGM} combines generative learning with spectral contrastive learning. HSP \cite{HSP} uses a hierarchical semantic preservation framework to integrate 2D and 3D skeleton information. Although these approaches provide useful alternatives, they often require iterative generation, additional modalities, or multi-stream inputs such as bone and motion modalities. In contrast, SLiM focuses on a single 3D joint stream and studies compact-token skeleton representation learning through decoder-free masked feature prediction.

\section{Method}
\label{sec:method}

\subsection{Overview and Compact Tokenization}
\begin{figure}[t]
  \centering
  \includegraphics[width=0.8\textwidth]{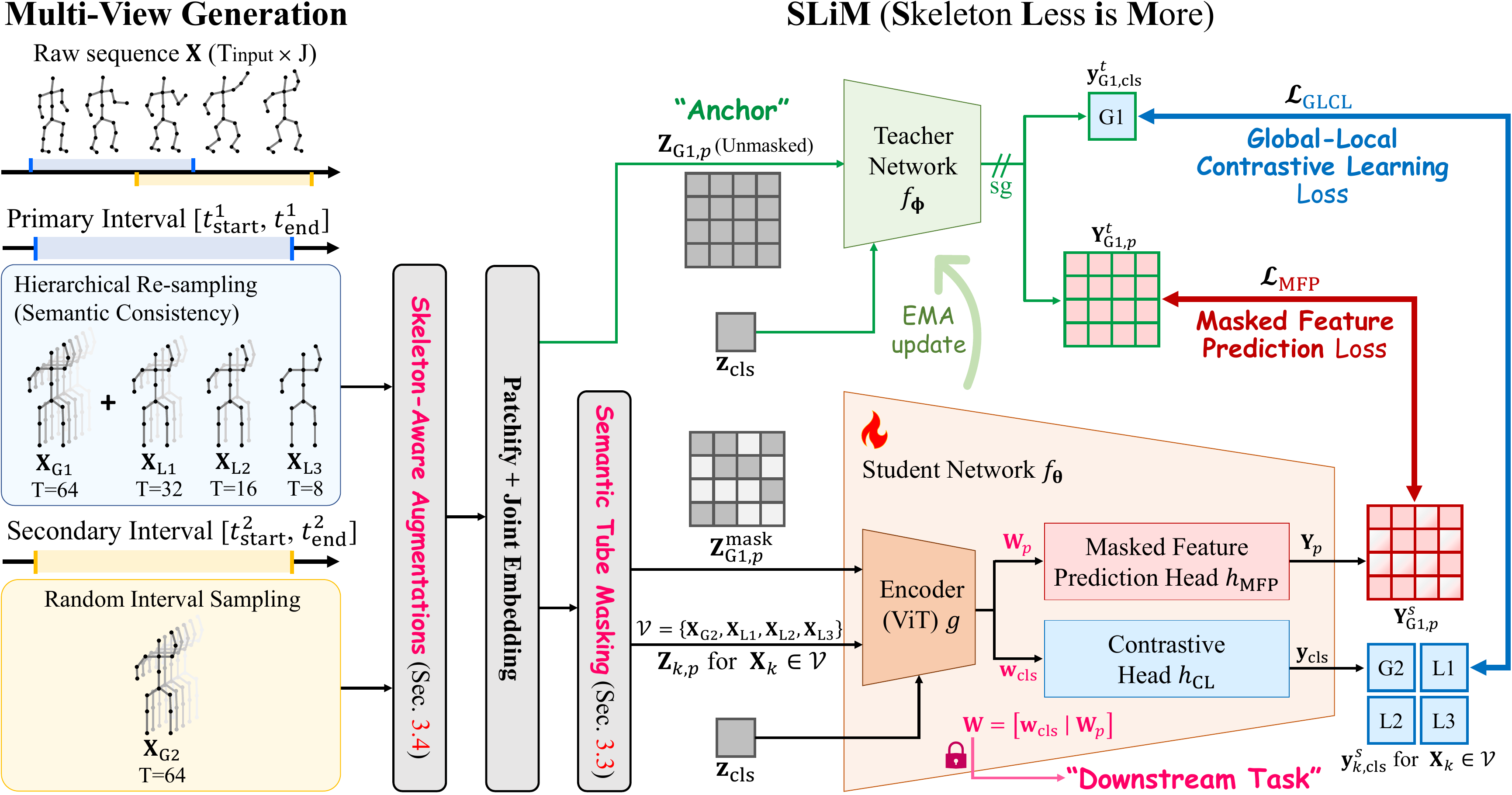}
  \caption{
  Overview of SLiM. SLiM combines decoder-free masked feature prediction (MFP) and global-local contrastive learning (GLCL) in a teacher--student framework. The student predicts teacher feature distributions from masked skeleton views and aligns global-local views for contrastive learning, using a shared compact-token encoder.}
  \label{fig:framework}
\end{figure}

Fig.~\ref{fig:framework} illustrates \textbf{SLiM} (\textbf{S}keleton \textbf{L}ess \textbf{i}s \textbf{M}ore), a compact-token skeleton SSL framework that combines masked feature prediction and contrastive learning in a shared encoder. SLiM follows a teacher--student distillation scheme \cite{Teacher}. The student network $f_{\bm\theta}$ is optimized by back-propagation, while the teacher network $f_{\bm\phi}$ is updated by an EMA of the student weights as $\bm\phi \leftarrow m \bm\phi + (1-m)\bm\theta$, where $m$ is a momentum coefficient. Both networks share the same architecture: an encoder $g$ and two lightweight projection heads $h_\mathrm{MFP}$ and $h_\mathrm{CL}$, for masked feature prediction and contrastive learning, respectively.

\noindent\textbf{Encoder and tokenization.}
We employ a Vision Transformer (ViT) \cite{ViT} as the backbone encoder $g$. Let the input skeleton clip be denoted as $\mathbf{X}\in \mathbb{R}^{T \times J \times C_\mathrm{in}}$, where $T$ is the sampled clip length, $J$ is the number of joints, and $C_\mathrm{in}=3$ is the coordinate dimension. We reshape $\mathbf{X}$ into skeletal-temporal patches $\mathbf{X}_p \in \mathbb{R}^{N \times D_p}$, where
\begin{equation}
    N = N_T N_J, \qquad
    N_T = \frac{T}{P_T}, \qquad
    N_J = \frac{J}{P_J}, \qquad
    D_p = P_T P_J C_\mathrm{in}.
\end{equation}
Here, $(P_T,P_J)$ denotes the temporal and joint patch size, and $N_T \times N_J$ is the encoder token grid after patchification, not the raw input length. Each patch token is linearly projected by $\mathbf{E}\in \mathbb{R}^{D_p \times D}$ and combined with learnable skeletal positional embeddings $\mathbf{E}_\mathrm{skel}\in \mathbb{R}^{N_J \times D}$:
\begin{equation}
    \mathbf{Z}_p =
    \left[
    \mathbf{x}_p^{1}\mathbf{E}
    \mid
    \mathbf{x}_p^{2}\mathbf{E}
    \mid
    \cdots
    \mid
    \mathbf{x}_p^{N}\mathbf{E}
    \right]
    + \mathbf{E}_\mathrm{skel}^{b},
    \label{eq:patch_embed}
\end{equation}
where $\mathbf{Z}_p\in\mathbb{R}^{N\times D}$ and $\mathbf{E}_\mathrm{skel}^{b}\in\mathbb{R}^{N\times D}$ is the broadcast version of $\mathbf{E}_\text{skel}$ along the temporal dimension. We append a learnable class token $\mathbf{z}_{\texttt{cls}}\in\mathbb{R}^{1\times D}$ to form $\mathbf{Z}_0=[\mathbf{z}_{\texttt{cls}}\mid \mathbf{Z}_p]$. To model temporal dependencies, we adopt 1D temporal rotary positional embeddings (RoPE) \cite{RoPE}. Specifically, RoPE is applied to the query and key vectors according to the temporal token index before computing self-attention, while joint identity is encoded by the skeletal positional embedding. Given the initialized token sequence $\mathbf{Z}_0$, the encoder $g$ produces
the final token representations:
\[
    \mathbf{W}=g(\mathbf{Z}_0)=[\mathbf{w}_{\texttt{cls}}\mid \mathbf{W}_p],
\]
where $\mathbf{W}$ is the final representation used for downstream evaluation.

\noindent\textbf{Projection heads.}
Given the encoder output $\mathbf{W}=[\mathbf{w}_{\texttt{cls}}\mid \mathbf{W}_p]$,
we use two projection heads for the global and patch-wise objectives. The contrastive head $h_{\mathrm{CL}}$ maps the class token to $K$-dimensional logits for global-local contrastive learning, while the masked feature prediction head $h_{\mathrm{MFP}}$ maps patch-wise features to $K$-dimensional logits for masked feature prediction:

\begin{equation}
    \mathbf{y}_{\texttt{cls}} = h_{\mathrm{CL}}(\mathbf{w}_{\texttt{cls}}) \in \mathbb{R}^{1 \times K},
    \qquad
    \mathbf{Y}_p = h_{\mathrm{MFP}}(\mathbf{W}_p) \in \mathbb{R}^{N \times K},
\end{equation}

where $\mathbf{y}_{\texttt{cls}}$ is used for the sequence-level contrastive
objective, and $\mathbf{Y}_p$ is used for masked feature prediction at the patch level.

\subsection{Decoder-Free Masked Feature Prediction with Contrastive Learning}
\label{sec:mfp_glcl}

\noindent\textbf{Multi-view generation.}
To learn representations across different temporal granularities, we generate two global views $\mathbf{X}_{\mathrm{G1}}$ and $\mathbf{X}_{\mathrm{G2}}$ by sampling temporal intervals from the raw skeleton sequence. Given the primary global view $\mathbf{X}_{\mathrm{G1}}$, we further generate local views $\mathbf{X}_{\mathrm{L1}}$, $\mathbf{X}_{\mathrm{L2}}$, and $\mathbf{X}_{\mathrm{L3}}$ by re-sampling from the same temporal interval at reduced resolutions of 32, 16, and 8 frames. For each resolution, we sample two independent local crops, resulting in six
local views in total: 
$\mathcal{V}_{\mathrm{local}}=
\{\mathbf{X}_{\mathrm{L1}}^{1},\mathbf{X}_{\mathrm{L1}}^{2},
\mathbf{X}_{\mathrm{L2}}^{1},\mathbf{X}_{\mathrm{L2}}^{2},
\mathbf{X}_{\mathrm{L3}}^{1},\mathbf{X}_{\mathrm{L3}}^{2}\}$. Anchoring local views to the same global interval reduces semantic misalignment between views and encourages temporal scale invariance. All views are processed by Skeleton-Aware Augmentations in Sec.~\ref{sec:aug}. We define the following losses using $\mathbf{X}_{\mathrm{G1}}$ as the anchor and apply the same procedure symmetrically with $\mathbf{X}_{\mathrm{G2}}$ as the anchor. 

\noindent\textbf{Masked feature prediction (MFP).}
We formulate masked modeling as feature prediction rather than coordinate reconstruction. The teacher network receives the unmasked global view $\mathbf{X}_{\mathrm{G1}}$ and produces patch-level target features:

\begin{equation}
    [\mathbf{y}_{\mathrm{G1},\texttt{cls}}^{t}\mid \mathbf{Y}_{\mathrm{G1},p}^{t}]
    =
    f_{\bm\phi}\!\left([\mathbf{z}_{\texttt{cls}}\mid \mathbf{Z}_{\mathrm{G1},p}]\right).
\end{equation}
For the student network, a binary mask $\mathcal{M}\in\{0,1\}^{N}$ is applied to the same patch tokens:
\begin{equation}
    \mathbf{Z}_{\mathrm{G1},p}^{\texttt{mask}}
    =
    (1-\mathcal{M})\odot \mathbf{Z}_{\mathrm{G1},p}
    +
    \mathcal{M}\odot \mathbf{e}_{\texttt{[mask]}}^{b},
    \label{eq:mask}
\end{equation}
where $\mathcal{M}_i=0$ denotes a visible token, $\mathcal{M}_i=1$ denotes a masked token, $\mathbf{e}_{\texttt{[mask]}}\in\mathbb{R}^{D}$ is a learnable mask token, and $\mathbf{e}_{\texttt{[mask]}}^{b}\in\mathbb{R}^{N\times D}$ is its broadcast version. The student network prediction is:
\begin{equation}
    [\mathbf{y}_{\mathrm{G1},\texttt{cls}}^{s}\mid \mathbf{Y}_{\mathrm{G1},p}^{s}]
    =
    f_{\bm\theta}\!\left([\mathbf{z}_{\texttt{cls}}\mid \mathbf{Z}_{\mathrm{G1},p}^{\texttt{mask}}]\right).
\end{equation}

Let $\Omega=\{i\mid \mathcal{M}_i=1\}$ be the set of masked patch tokens. For each masked token $i$, we obtain $K$-dimensional probability distributions by applying softmax to the teacher and student patch outputs:
\begin{equation}
    \mathbf{p}_{i}^{t}=\mathrm{softmax}\left(\mathbf{Y}_{i}^{t}/\tau_t\right),
    \qquad
    \mathbf{p}_{i}^{s}=\mathrm{softmax}\left(\mathbf{Y}_{i}^{s}/\tau_s\right),
    \label{eq:prob}
\end{equation}
where $\tau_t$ and $\tau_s$ are temperature parameters. We adopt the teacher-student cross-entropy loss \cite{iBOT} and apply it to masked skeleton tokens:

\begin{equation}
    \mathcal{L}_{\mathrm{MFP}}
    =
    -\frac{1}{|\Omega|}
    \sum_{i\in\Omega}
    \sum_{c=1}^{K}
    \mathrm{sg}\!\left(\mathbf{p}_{i}^{t}(c)\right)
    \log \mathbf{p}_{i}^{s}(c),
\end{equation}

where $\mathrm{sg}(\cdot)$ denotes stop-gradient. Since MFP predicts teacher features rather than raw coordinates, SLiM removes the reconstruction decoder during pre-training. The downstream inference cost is then governed by the compact encoder token grid used for evaluation.

\noindent\textbf{Global-local contrastive learning (GLCL).}
We use the teacher network's global class token from $\mathbf{X}_{\mathrm{G1}}$ as the anchor and align it with student network predictions from the other global and local views. Let
$\mathcal{V}=\{X_{\mathrm{G2}}\}\cup \mathcal{V}_{\mathrm{local}}$.
For each $\mathbf{X}_k\in\mathcal{V}$, the student produces:
\begin{equation}
    [\mathbf{y}_{k,\texttt{cls}}^{s}\mid \mathbf{Y}_{k,p}^{s}]
    =
    f_{\bm\theta}\!\left([\mathbf{z}_{\texttt{cls}}\mid \mathbf{Z}_{k,p}]\right).
\end{equation}
Note that local views are sampled from the same temporal interval as $\mathbf{X}_{\mathrm{G1}}$. The GLCL loss is:
\begin{equation}
    \mathcal{L}_{\mathrm{GLCL}}
    =
    -\frac{1}{|\mathcal{V}|}
    \sum_{\mathbf{X}_k\in\mathcal{V}}
    \sum_{c=1}^{K}
    \mathrm{sg}\!\left(\mathbf{p}_{\mathrm{G1},\texttt{cls}}^{t}(c)\right)
    \log \mathbf{p}_{k,\texttt{cls}}^{s}(c),
    \label{eq:glcl}
\end{equation}
where $\mathbf{p}_{\mathrm{G1},\texttt{cls}}^{t}$ and $\mathbf{p}_{k,\texttt{cls}}^{s}$ are the $K$-dimensional probability distributions obtained in the same manner as Eq.~\ref{eq:prob}. This encourages temporal scale invariance by aligning local clips with the global semantic context provided by the teacher network.

\noindent\textbf{Total objective.}
SLiM jointly optimizes masked feature prediction and global-local contrastive learning as
$\mathcal{L}_{\mathrm{total}}=\mathcal{L}_{\mathrm{MFP}}+\lambda\mathcal{L}_{\mathrm{GLCL}}$,
where we empirically set $\lambda=1.0$ for all experiments.

\subsection{Semantic Tube Masking (STM)}
\label{sec:mask}

\begin{figure}[t]
  \centering
  \includegraphics[width=0.8\linewidth]{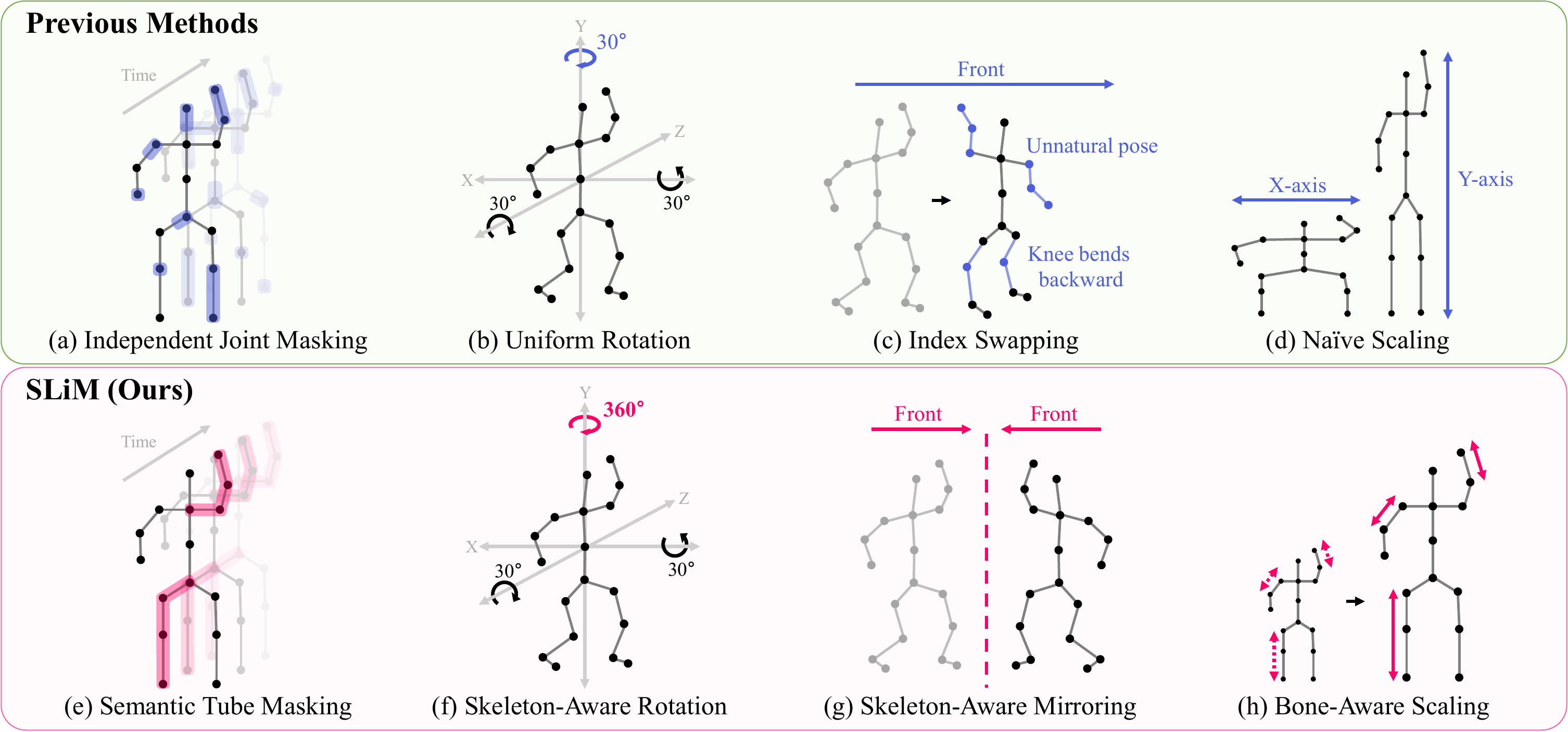}
  \caption{
  Comparison of masking and augmentation strategies.
  Top: independent joint masking and conventional geometric augmentations may allow local interpolation shortcuts or distort skeleton structure.
  Bottom: Semantic Tube Masking masks connected anatomical regions over time, while Skeleton-Aware Augmentations preserve basic anatomical consistency during view generation.
  }
  \label{fig:augmentation}
  \vspace{-0.2cm}
\end{figure}

Independent joint- or patch-level masking can encourage shortcut solutions, since skeleton joints are strongly correlated in space and time. As illustrated in Fig.~\ref{fig:augmentation}(a), independently masking isolated joints can leave nearby spatial and temporal cues that allow local interpolation. We propose Semantic Tube Masking (STM), shown in Fig.~\ref{fig:augmentation}(e), which masks connected anatomical regions, such as an arm, leg, or torso, over consecutive frames. Compared to isolated joint masking, STM occludes entire functional body parts across time, forcing the model to infer missing features from broader body context and motion dynamics.

\noindent\textbf{Skeletal-temporal tube design.}
We define semantic joint groups according to the human body structure, such as torso, arms, and legs. During mask generation, STM first selects an anatomical group and then samples a connected subset of joints within that group. The selected joints are masked over a consecutive temporal span, forming a skeletal-temporal tube. To balance masking difficulty across body parts, we adopt a constant-volume strategy: the temporal duration is adjusted inversely to the spatial size of the selected joint subset. Thus, smaller regions are masked for longer temporal spans, while larger regions are masked for shorter spans. This design reduces reliance on local interpolation and encourages prediction from broader skeletal-temporal context.

\noindent\textbf{Dual role in training.}
STM is used in two ways. For MFP, it defines the masked patch tokens replaced by $\mathbf{e}_{\texttt{[mask]}}$ in Eq.~\ref{eq:mask}. For GLCL, it is applied stochastically to local views as a structural augmentation, encouraging the representation to remain stable under partial anatomical occlusion.

\subsection{Skeleton-Aware Augmentations (SAA)}
\label{sec:aug}

Standard geometric augmentations \cite{ActCLR, AimCLR, CrosSCLR} may disregard the articulated structure of the human body, leading to distorted skeletons or unreliable contrastive pairs, which can harm representation learning, as illustrated in Fig.~\ref{fig:augmentation}(b--d). We introduce Skeleton-Aware Augmentations (SAA), shown in Fig.~\ref{fig:augmentation}(f--h), a set of transformations designed to improve view diversity while preserving basic anatomical consistency.

\noindent \textbf{Skeleton-aware rotation.} This is illustrated in Fig.~\ref{fig:augmentation}(f) and compared with the uniform rotation in Fig.~\ref{fig:augmentation}(b). Standard rotation \cite{ActCLR, CTRGCN, AimCLR} is typically confined to a limited range of angles. To maintain a realistic upright posture, we apply a full $360^\circ$ random rotation exclusively around the vertical $Y$-axis. For the remaining non-gravity axes ($X, Z$), we constrain perturbations to small angles (e.g., $30^\circ$) to avoid physically implausible tilting. Unlike previous works that restrict all axes to maintain stability, our approach decouples the rotation logic to maximize view diversity while preserving physical validity.

\noindent \textbf{Skeleton-aware mirroring.} As shown in Fig.~\ref{fig:augmentation}(g), and compared with the index swapping in Fig.~\ref{fig:augmentation}(c), heuristic index swapping \cite{ActCLR, AimCLR, Skateformer} is a common approach but fails to achieve a mathematically rigorous reflection, thereby resulting in unnatural poses. Our strategy performs a geometric mirroring that simultaneously negates coordinate values along the lateral axis ($X$-axis) and reassigns joint indices according to bilateral symmetry. This dual-operation ensures that the resulting motion is anatomically consistent--preserving the correct ``Front'' orientation and left-right semantics--rather than producing a distorted artifact.

\noindent \textbf{Bone-aware scaling.} This is depicted in Fig.~\ref{fig:augmentation}(h) and compared with the naive scaling in Fig.~\ref{fig:augmentation}(d). The naive scaling of joint coordinates \cite{AimCLR, CTRGCN} can disrupt the structure of the skeleton. To preserve the essence of the action, we decompose each bone into a directional unit vector $\hat{\mathbf{v}}$ and a scalar length $l$. We apply random scaling factors only to $l$ while strictly maintaining $\hat{\mathbf{v}}$. This effectively simulates the same action performed by subjects of varying body sizes and proportions, thereby enriching the intra-class variance without semantic distortion.
\section{Experimental Results}
\label{sec:exp}

\subsection{Datasets and Protocols}
\label{sec:dataset}
We evaluate SLiM on three standard skeleton-based action recognition benchmarks:
NTU RGB+D 60 (NTU-60) \cite{NTU60}, NTU RGB+D 120 (NTU-120) \cite{NTU120},
and PKU-MMD II \cite{PKU}. NTU-60 contains 56,880 skeleton sequences from
60 action classes and provides Cross-Subject (X-Sub) and Cross-View (X-View)
protocols. NTU-120 extends NTU-60 to 120 classes and 114,480 samples, with
Cross-Subject (X-Sub) and Cross-Setup (X-Set) protocols. PKU-MMD II contains
51 action classes, and we follow the standard Cross-Subject (X-Sub) protocol.
Unless otherwise specified, all experiments use only the 3D joint modality,
without multi-stream fusion from bone or motion modalities.

\begin{table}[tbp]
    \scriptsize
    \centering
    \caption{
    Linear evaluation on NTU-60, NTU-120, and PKU-MMD II. 
    We report Top-1 accuracy (\%) using the joint modality.
    \textbf{Bold} and \underline{underline} indicate the best and second-best results, respectively.
    }
    \label{tab:action_recognition}
    \resizebox{0.8\textwidth}{!}{%
    \def\arraystretch{1.0}
    \begin{tabular}{llccccc}
    \toprule
    \multirow{2}{*}{\textbf{Method}} & \multirow{2}{*}{\textbf{Publication}} 
    & \multicolumn{2}{c}{\textbf{NTU-60}} 
    & \multicolumn{2}{c}{\textbf{NTU-120}} 
    & \textbf{PKU-MMD II} \\
    \cmidrule(lr){3-4} \cmidrule(lr){5-6} \cmidrule(lr){7-7}
    & & X-Sub & X-View & X-Sub & X-Set & X-Sub \\
    \midrule
    \multicolumn{7}{l}{\textit{Other skeleton SSL baselines:}}\\
    GL-Transformer \cite{GL-Transformer} & ECCV'22 & 76.3 & 83.8 & 66.0 & 68.7 & -- \\
    MacDiff \cite{MacDiff} & ECCV'24 & 86.4 & 91.0 & 79.4 & 80.2 & -- \\
    IGM \cite{IGM} & ECCV'24 & 86.2 & 91.2 & 80.0 & 81.4 & -- \\
    USDRL \cite{USDRL} & AAAI'25 & 85.2 & 91.7 & 76.6 & 78.1 & 54.4 \\
    HSP \cite{HSP} & CVPR'25 & 80.7 & 88.0 & 71.0 & 73.2 & 48.9 \\
    \midrule
    \multicolumn{7}{l}{\textit{Contrastive learning (CL):}}\\
    CPM \cite{CPM} & ECCV'22 & 78.7 & 84.9 & 68.7 & 69.6 & 48.3 \\
    CMD \cite{CMD} & ECCV'22 & 79.8 & 86.9 & 70.3 & 71.5 & 43.0 \\
    AimCLR \cite{AimCLR} & AAAI'22 & 74.3 & 79.7 & 63.4 & 63.4 & -- \\
    RVTCLR \cite{RVTCLR} & ICCV'23 & 74.7 & 79.1 & -- & -- & -- \\
    PTSL \cite{PTSL} & AAAI'23 & 77.3 & 81.8 & 66.2 & 67.7 & 49.3 \\
    HYSP \cite{HYSP} & ICLR'23 & 78.2 & 82.6 & 61.8 & 64.6 & -- \\
    HaLP \cite{HaLP} & CVPR'23 & 79.7 & 86.8 & 71.1 & 72.2 & 43.5 \\
    ActCLR \cite{ActCLR} & CVPR'23 & 80.9 & 86.7 & 69.0 & 70.5 & -- \\
    \midrule
    \multicolumn{7}{l}{\textit{Masked / predictive modeling:}}\\
    SkeletonMAE \cite{SkeletonMAE} & ICMEW'23 & 74.8 & 77.7 & 72.5 & 73.5 & 36.1 \\
    MAMP \cite{MAMP} & ICCV'23 & 84.9 & 89.1 & 78.6 & 79.1 & 53.8 \\
    S-JEPA \cite{SJEPA} & ECCV'24 & 85.3 & 89.8 & 79.6 & 79.9 & 53.5 \\
    GFP \cite{GFP} & ICCV'25 & 85.9 & 92.0 & 79.1 & 80.3 & 56.2 \\
    AMR \cite{AMR} & CVPR'26 & \underline{87.4} & \underline{92.3}
    & \underline{81.1} & \underline{81.9} & \textbf{60.3} \\
    \midrule
    \rowcolor{orange!12}
    \multicolumn{7}{l}{\textit{Unified MFP + CL:}}\\
    \rowcolor{orange!12}
    \textbf{SLiM (Ours)} & -- & \textbf{87.9} & \textbf{93.2} 
    & \textbf{81.2} & \textbf{83.6} & \underline{59.7} \\
    \bottomrule
    \end{tabular}}
\end{table}

\subsection{Implementation Details}
\label{sec:implementation}

\noindent\textbf{Architecture.}
Following previous MAE-based skeleton SSL methods \cite{SkeletonMAE, MAMP, SJEPA, GFP},
we use a ViT \cite{ViT} encoder with 8 Transformer layers, hidden dimension
$D=256$, and 8 attention heads. This matches the encoder depth and width of
existing baselines, while SLiM uses a compact token grid. Specifically, global
views are sampled to $T=64$ frames with $J=25$ joints, and patchified with
$P_T=8$ and $P_J=1$, resulting in an encoder token grid of
$N_T \times N_J = 8 \times 25$.

\noindent\textbf{Pre-training setup.}
SLiM is pre-trained for 150 epochs with the teacher--student framework.
The teacher momentum $m$ is increased from 0.994 to 1.0 during training.
We use AdamW \cite{AdamW} with a base learning rate of $2.0 \times 10^{-4}$,
a 20-epoch linear warmup, and a cosine decay \cite{CosineAnneal} to
$1.0 \times 10^{-6}$. The model is trained with a total batch size of 768
on 4 NVIDIA RTX A6000 GPUs.

\noindent\textbf{Masking and augmentations.}
For Semantic Tube Masking (STM), the target mask ratio is sampled uniformly
from $[0.5, 0.9]$. For masked feature prediction, STM is applied to the primary
global view by replacing selected patch tokens with learnable mask tokens.
For global-local contrastive learning, STM is applied stochastically to local
views as a structural perturbation. For Skeleton-Aware Augmentations (SAA),
each transformation is applied independently with probability $0.5$. Rotation
uses a full $360^\circ$ range around the vertical $Y$-axis and limits the
non-gravity axes $(X,Z)$ to $\pm 30^\circ$. Bone-aware scaling samples length
scales from $[0.85,1.15]$ while preserving bone direction vectors.

\subsection{Performance Comparison}

We compare SLiM with SOTA skeleton SSL methods under three protocols: linear evaluation, semi-supervised learning, and action retrieval.
To isolate representation quality, we report results using the single joint
modality and exclude gains from multi-stream fusion such as bone or motion streams.

\noindent\textbf{Linear evaluation.}
Table~\ref{tab:action_recognition} reports Top-1 accuracy under the standard
linear evaluation protocol, where the pre-trained encoder is frozen and only a
linear classifier is trained. Including concurrent AMR \cite{AMR}, SLiM obtains
the best result on four of five protocols, exceeding AMR by 0.5, 0.9, 0.1,
and 1.7\%-points on NTU-60 X-Sub, NTU-60 X-View, NTU-120 X-Sub, and NTU-120
X-Set, respectively. On PKU-MMD II, SLiM ranks second, 0.6\%-points below AMR.
Importantly, these results are obtained with a compact encoder token grid.
As summarized in Fig.~\ref{fig:token}, coordinate-reconstruction MAE baselines
degrade noticeably when the temporal token budget is reduced, whereas SLiM
maintains strong accuracy under compact token budgets. Table~\ref{tab:flops}
further provides the computational breakdown: SLiM uses an $8\times25$ encoder
token grid and requires only 3.59 GFLOPs for downstream inference, yielding a
7.89$\times$ reduction compared to dense-token MAE-based methods with
$30\times25$ tokens. During pre-training, SLiM also removes the decoder, reducing the computation spent on components that are
discarded after pre-training.

\begin{table}[t]
\centering
\scriptsize
\caption{
Additional downstream evaluations on NTU-60.
(a) Semi-supervised action recognition using 1\% and 10\% labeled data.
(b) Action retrieval using $k$-NN with $k=1$.
}
\label{tab:semi_retrieval}
\begin{minipage}[t]{0.47\textwidth}
    \centering
    {\small (a) Semi-supervised learning}\\[2pt]
    \resizebox{\textwidth}{!}{%
    \def\arraystretch{1.0}
    \begin{tabular}{lcccc}
        \toprule
        \multirow{2}{*}{\textbf{Method}} & \multicolumn{2}{c}{\textbf{X-Sub}} 
        & \multicolumn{2}{c}{\textbf{X-View}} \\
        \cmidrule(lr){2-3} \cmidrule(lr){4-5}
        & 1\% & 10\% & 1\% & 10\% \\
        \midrule
        CPM \cite{CPM} & 56.7 & 73.0 & 57.5 & 77.1 \\
        CMD \cite{CMD} & 50.6 & 75.4 & 53.0 & 80.2 \\
        HYSP \cite{HYSP} & -- & 76.2 & -- & 80.4 \\
        HaLP \cite{HaLP} & 46.6 & 72.6 & 48.7 & 77.1 \\
        HiCo \cite{HiCo} & 54.4 & 73.0 & 54.8 & 78.3 \\
        USDRL \cite{USDRL} & 57.3 & 80.2 & 60.7 & 84.0 \\
        \midrule
        SkeletonMAE \cite{SkeletonMAE} & 54.4 & 80.6 & 54.6 & 83.5 \\
        MAMP \cite{MAMP} & 66.0 & 88.0 & 68.7 & 91.5 \\
        S-JEPA \cite{SJEPA} & 67.5 & 88.4 & 69.1 & 91.4 \\
        GFP \cite{GFP} & \underline{71.8} & \underline{88.7} 
        & \underline{72.9} & \textbf{92.1} \\
        \midrule
        \rowcolor{orange!12}
        \textbf{SLiM (Ours)} & \textbf{72.1} & \textbf{88.8} 
        & \textbf{75.8} & \underline{91.9} \\
        \bottomrule
    \end{tabular}}
\end{minipage}
\hfill
\begin{minipage}[t]{0.38\textwidth}
    \centering
    {\small (b) Action retrieval}\\[2pt]
    \resizebox{\textwidth}{!}{%
    \def\arraystretch{1.0}
    \begin{tabular}{lcc}
        \toprule
        \multirow{2}{*}{\textbf{Method}} & \multicolumn{2}{c}{\textbf{NTU-60}} \\
        \cmidrule(lr){2-3}
         & X-Sub & X-View \\
        \midrule
        LongT-GAN \cite{LongTGAN} & 39.1 & 48.1 \\
        P\&C \cite{PandC} & 50.7 & 76.3 \\
        ISC \cite{ISC} & 62.5 & 82.6 \\
        HaLP \cite{HaLP} & 65.8 & 83.6 \\
        HiCo \cite{HiCo} & 68.3 & 84.8 \\
        SkeAttnCLR \cite{SkeAttnCLR} & 69.4 & 76.8 \\
        UmURL \cite{UmURL} & \underline{71.3} & \underline{88.3} \\
        \midrule
        MAMP \cite{MAMP} & 62.0 & 70.0 \\
        GFP \cite{GFP} & 70.9 & 87.1 \\
        \midrule
        \rowcolor{orange!12}
        \textbf{SLiM (Ours)} & \textbf{72.5} & \textbf{89.9} \\
        \bottomrule
    \end{tabular}}
\end{minipage}
\end{table}

\noindent\textbf{Semi-supervised learning and retrieval.}
Table~\ref{tab:semi_retrieval}(a) evaluates the label efficiency of the learned
representations. SLiM performs particularly well in the 1\% label setting,
improving the previous best by 0.3 and 2.9\%-points on X-Sub and
X-View, respectively, while remaining competitive in the 10\% setting.
Table~\ref{tab:semi_retrieval}(b) further evaluates frozen representations
with non-parametric retrieval. SLiM achieves the best retrieval accuracy on both
protocols, indicating that the learned features are discriminative even without
training a task-specific classifier.

\begin{table}[t]
    \centering
    \scriptsize
    \begin{minipage}[t]{0.57\textwidth}
        \centering
        \caption{
        Token budget and computational cost comparison on NTU-60.
        \(N_T\times N_J\) denotes the encoder token grid after patchification. SLiM reduces encoder inference cost by 7.89$\times$ compared to dense-token MAE-based methods.
        }
        \label{tab:flops}
        \resizebox{\textwidth}{!}{%
        \begin{tabular}{lcccc}
        \toprule
        \multirow{2}{*}{\textbf{Methods}} & \textbf{Tokens} & \makecell{\textbf{Inference}\\[1.5pt](GFLOPs)} & \multicolumn{2}{c}{\makecell{\textbf{Training}\\[1.5pt](GFLOPs)}} \\
        \cmidrule(lr){2-2} \cmidrule(lr){3-3} \cmidrule(lr){4-5}
        & $N_T \times N_J$ & Encoder & Encoder & Decoder \\
        \midrule
        SkeletonMAE \cite{SkeletonMAE} & $30 \times 25$ & 28.32 & 1.97 & 17.70 \\
        MAMP \cite{MAMP} & $30 \times 25$ & 28.32 & 1.97 & 17.70 \\
        S-JEPA \cite{SJEPA} & $30 \times 25$ & 28.32 & 1.97 & 17.70 \\
        GFP \cite{GFP} & $30 \times 25$ & 28.32 & 1.97 & 1.57 \\
        \rowcolor{orange!15}
        \textbf{SLiM (Ours)} & $8 \times 25$ & \textbf{3.59} & 3.59 & -- \\
        \bottomrule
    \end{tabular}}
    \end{minipage}
    \hfill
    \begin{minipage}[t]{0.40\textwidth}
        \centering
        \caption{Ablation studies for the objective functions on NTU-60. $\mathcal{L}_{\text{CL}}$ represents a standard CL loss without temporal diversity.}
        \label{tab:ablation_loss}
        \resizebox{\textwidth}{!}{%
        \def\arraystretch{1.0}
        \begin{tabular}{c|cc|cc}
            \toprule
            \multicolumn{3}{c|}{\textbf{Objective}} & \multicolumn{2}{c}{\textbf{NTU-60}} \\
            \midrule
            $\mathcal{L}_{\mathrm{MFP}}$ & $\mathcal{L}_{\mathrm{CL}}$ & $\mathcal{L}_{\mathrm{GLCL}}$ & X-Sub & X-View \\
            \midrule
            \ding{55} & \ding{51} & \ding{55} & 73.6 & 78.9 \\
            \ding{55} & \ding{55} & \ding{51} & 75.3 & 80.1 \\
            \ding{51} & \ding{55} & \ding{55} & 85.3 & 90.6 \\
            \ding{51} & \ding{51} & \ding{55} & 86.2 & 91.3 \\
            \rowcolor{orange!15}
            \ding{51} & \ding{55} & \ding{51} & 87.7 & 92.7 \\
            \bottomrule
        \end{tabular}}
    \end{minipage}
\end{table}

\subsection{Ablation Studies}

We conduct ablation studies on NTU-60. For efficiency, all ablation models are
pre-trained for 100 epochs, while the final model is pre-trained for 150 epochs.
Thus, the ablation results may be slightly lower than the final model reported
in Table~\ref{tab:action_recognition}. In the following analyses, ``Limited''
denotes replacing the proposed component with its conventional counterpart
illustrated in Fig.~\ref{fig:augmentation}(a--d).

\noindent\textbf{Objective functions.}
Table~\ref{tab:ablation_loss} evaluates the contribution of masked feature
prediction and contrastive learning. $\mathcal{L}_{\text{CL}}$-only and $\mathcal{L}_{\text{GLCL}}$-only variants show limited
performance, indicating that discriminative losses alone are insufficient
for learning detailed skeleton representations. $\mathcal{L}_{\text{MFP}}$ alone provides a strong
baseline by learning local predictive features, achieving 85.3\% on X-Sub.
Adding a standard CL loss improves the result to 86.2\%, while replacing it with
$\mathcal{L}_{\text{GLCL}}$ further improves performance to 87.7\%. This suggests that $\mathcal{L}_{\text{MFP}}$ and
temporally diverse global-local contrastive learning are complementary: $\mathcal{L}_{\text{MFP}}$ captures local skeletal-temporal patterns, while $\mathcal{L}_{\text{GLCL}}$ improves sequence-level
semantic discrimination across temporal granularities.

\begin{table}[t]
\centering
\scriptsize
\caption{
Ablation studies on NTU-60.
(a) Dual-role Semantic Tube Masking (STM) for the global masked-view branch and
local contrastive-view branch. 
(b) Skeleton-Aware Augmentations (SAA), including rotation, mirroring, and scaling.
Limited denotes the corresponding conventional masking or augmentation strategy.
}
\label{tab:ablation_stm_saa}

\begin{minipage}[t]{0.37\textwidth}
    \centering
    {\small (a) Semantic Tube Masking}\\[2pt]
    \resizebox{\textwidth}{!}{%
    \def\arraystretch{1.0}
    \setlength{\tabcolsep}{7pt}
    \begin{tabular}{cc|cc}
        \toprule
        \multicolumn{2}{c|}{\textbf{Masking}} & \multicolumn{2}{c}{\textbf{NTU-60}} \\
        \midrule
        Global View & Local View & X-Sub & X-View \\
        \midrule
        \ding{55} & \ding{51} & 75.3 & 80.1 \\
        \ding{51} (Limited) & \ding{51} & 84.2 & 90.4 \\
        \ding{51} & \ding{55} & 83.3 & 89.4 \\
        \ding{51} & \ding{51} (Limited) & 85.7 & 89.9 \\
        \midrule
        \ding{51} (90\%) & \ding{51} (90\%) & 86.8 & 92.1 \\
        \ding{51} (0--50\%) & \ding{51} (0--50\%) & 84.5 & 89.6 \\
        \rowcolor{orange!12}
        \ding{51} (50--90\%) & \ding{51} (50--90\%) & 87.7 & 92.7 \\
        \bottomrule
    \end{tabular}}
\end{minipage}
\hfill
\begin{minipage}[t]{0.50\textwidth}
    \centering
    {\small (b) Skeleton-Aware Augmentations}\\[2pt]
    \resizebox{\textwidth}{!}{%
    \def\arraystretch{1.0}
    \setlength{\tabcolsep}{7pt}
    \begin{tabular}{ccc|cc}
        \toprule
        \textbf{Rotation} & \textbf{Mirroring} & \textbf{Scaling} 
        & \textbf{X-Sub} & \textbf{X-View} \\
        \midrule
        \ding{55} & \ding{51} & \ding{51} & 81.6 & 86.9 \\
        \ding{51} (Limited) & \ding{51} & \ding{51} & 84.8 & 89.9 \\
        \ding{51} & \ding{55} & \ding{51} & 85.3 & 90.6 \\
        \ding{51} & \ding{51} (Limited) & \ding{51} & 85.9 & 91.0 \\
        \ding{51} & \ding{51} & \ding{55} & 84.4 & 90.0 \\
        \ding{51} & \ding{51} & \ding{51} (Limited) & 86.4 & 91.7 \\ 
        \rowcolor{orange!12}
        \ding{51} & \ding{51} & \ding{51} & 87.7 & 92.7 \\
        \bottomrule
    \end{tabular}}
\end{minipage}
\end{table}

\textbf{Semantic Tube Masking (STM).}
Table~\ref{tab:ablation_stm_saa}(a) evaluates STM's dual role in generating masked global views (MFP) and perturbing local views (GLCL). Replacing STM with conventional joint masking \cite{MAMP, SJEPA, GFP} in either role degrades X-Sub accuracy by up to 3.5\%-point (87.7\% vs. 84.2\%), confirming that structured anatomical masking is crucial to prevent trivial interpolation shortcuts. Additionally, our stochastic masking strategy (50--90\%) outperforms fixed extreme masking (90\%) by 0.9\%-point and low-ratio masking (0--50\%) by 3.2\%-point on X-Sub.

\textbf{Skeleton-Aware Augmentations (SAA).}
Table~\ref{tab:ablation_stm_saa}(b) isolates the contributions of geometric augmentations. By substituting standard augmentations from prior works \cite{ActCLR, AimCLR} with our SAA, we strictly preserve the 3D geometric validity of contrastive pairs. This anatomical consistency directly translates to enhanced downstream accuracy. Consequently, compared to their conventional counterparts, our rotation, mirroring, and scaling yield substantial improvements.

\section{Conclusion}
\label{sec:conclusion}

We presented \textbf{SLiM}, a compact-token skeleton SSL framework that replaces
coordinate reconstruction with decoder-free masked feature prediction and
combines it with global-local contrastive learning. Semantic Tube Masking and
Skeleton-Aware Augmentations help avoid interpolation shortcuts and provide reliable
views for skeleton representation learning. Across linear evaluation,
semi-supervised learning, and retrieval, SLiM achieves state-of-the-art performance while
reducing encoder inference computation by 7.89$\times$ compared to dense-token
MAE-based methods. Building upon these promising results, future work will explore extending our compact-token paradigm to more challenging scenarios, such as noisy 2D poses, missing joints, and heterogeneous skeleton topologies.

\section*{Acknowledgments}

This work was supported by IITP grant funded by the
Korea government (MSIT) (RS2022-00144444, Deep Learning Based Visual
Representational Learning and Rendering of Static and Dynamic Scenes).


\clearpage
\appendix

\section*{Appendix Overview}

\begin{table}[ht]
    \scriptsize
    \centering
    \caption{Overview of appendix contents.}
    \resizebox{0.95\columnwidth}{!}{
    \begin{tabular}{ll}
        \toprule
        \rowcolor{orange!15}
        \textbf{Section} & \textbf{Descriptions for Analysis and Discussion} \\
        \midrule
        \multirow{2}{*}{Sec.~\ref{sec:sup_A}} & \textbf{Implementation Details}:\\
        & Architecture, experiment settings, and hyper-parameters \\
        \midrule
        \multirow{2}{*}{Sec.~\ref{sec:sup_B}} & \textbf{Hierarchical Temporal Sampling}: \\
        & Global/local view generation \\
        \midrule
        Sec.~\ref{sec:stm} & \textbf{Semantic Tube Masking (STM)} (Algorithm~\ref{alg:stm})\\
        \midrule
        \multirow{2}{*}{Sec.~\ref{sec:saa}} & \textbf{Skeleton-Aware Augmentations (SAA)}:\\
        & Rotation (Algorithm~\ref{alg:saa_rot}), mirroring (Algorithm~\ref{alg:saa_mirror}), and scaling (Algorithm~\ref{alg:saa_scale}) \\
        \midrule
        \multirow{4}{*}{Sec.~\ref{sec:sup_E}} & \textbf{Additional Results and Discussion}: \\
        & Token-Budget Study (Table~\ref{tab:token_budget_avg}), detailed cost comparison (Table~\ref{tab:flops_supp}),\\
        & transfer learning (Table~\ref{tab:transfer}), and controlled prediction-target analysis (Table~\ref{tab:target_control}),\\
        & missing/noisy-joint robustness and degraded pre-training (Tables~\ref{tab:test_corruption}--\ref{tab:degraded_pretraining})\\
        \midrule
        Sec.~\ref{sec:limitations} & \textbf{Limitations} \\
        \midrule
        Sec.~\ref{sec:broader} & \textbf{Broader impact} \\
        \midrule
        Sec.~\ref{sec:assets} & \textbf{Compute resources and existing assets} \\
        \bottomrule
    \end{tabular}}
   \label{tab:supple}
\end{table}

\noindent
This appendix provides additional implementation details and analysis for SLiM.
Sec.~\ref{sec:sup_A} describes the architecture, projection heads, losses, and
hyperparameters. Sec.~\ref{sec:sup_B} details the hierarchical temporal sampling
strategy used for global and local view generation. Sec.~\ref{sec:stm} and
Sec.~\ref{sec:saa} provide the algorithmic details of Semantic Tube Masking and
Skeleton-Aware Augmentations, respectively. Sec.~\ref{sec:sup_E} reports
additional token-budget, cost, transfer, controlled-target, and robustness
analyses. Finally, Secs.~\ref{sec:limitations}--\ref{sec:assets}
discuss limitations, broader impact, compute resources, and existing assets.

\section{Implementation Details}
\label{sec:sup_A}

\noindent\textbf{Detailed architecture.}
Our encoder $g$ follows a ViT~\cite{ViT} backbone adapted to skeleton sequences,
with 8 Transformer layers, hidden dimension $D=256$, and 8 attention heads. The encoder is defined as:
\begin{equation}
\begin{aligned}
    \mathbf{Z}_0 &= [\mathbf{z}_{\texttt{cls}}\mid \mathbf{Z}_p], \\
    \mathbf{Z}_l^{\prime} &= \mathrm{MSA}_{\mathrm{RoPE}}\!\left(\mathrm{LN}\left(\mathbf{Z}_{l-1}\right)\right) + \mathbf{Z}_{l-1}, \\
    \mathbf{Z}_l &= \mathrm{MLP}\!\left(\mathrm{LN}\left(\mathbf{Z}_{l}^{\prime}\right)\right) + \mathbf{Z}_{l}^{\prime}, \qquad l=1,\ldots,L,\\
    \mathbf{W} &= [\mathbf{w}_{\texttt{cls}}\mid \mathbf{W}_p] = \mathrm{LN}\left(\mathbf{Z}_{L}\right),
\end{aligned}
\label{eq:encoder}
\end{equation}
where $\mathrm{MSA}_{\mathrm{RoPE}}$ denotes multi-head self-attention with temporal RoPE, $\mathrm{LN}$ denotes layer normalization, and $\mathbf{W}$ is the final representation used for downstream evaluation. Both the student network $f_{\bm\theta}$ and the teacher network $f_{\bm\phi}$ use the same encoder architecture and two
3-layer MLP projection heads, $h_{\mathrm{CL}}$ and $h_{\mathrm{MFP}}$.
Each head has a hidden dimension of 2,048 and a bottleneck dimension of 256,
and outputs $K=65{,}536$-dimensional prototype-assignment logits. The resulting
probability distributions used in the MFP and GLCL losses are therefore defined
over $K$ prototypes. The total objective balances the masked feature prediction loss
$\mathcal{L}_{\mathrm{MFP}}$ and the global-local contrastive loss
$\mathcal{L}_{\mathrm{GLCL}}$ with $\lambda=1.0$. 

\noindent\textbf{Settings for the token-budget study.}
For the token-budget study in Fig.~\ref{fig:token}, we vary the temporal token
number \(N_T\) by adjusting the sampled clip length and/or temporal patch size
while keeping the encoder depth and width fixed. Unless otherwise stated, all
variants are trained with the same optimization settings as their corresponding
baseline.

\noindent\textbf{Downstream evaluation details.}
For linear evaluation, we freeze the pre-trained encoder and train a single
linear classifier for 10 epochs using AdamW with learning rate $1 \times 10^{-4}$, and batch size 256. For semi-supervised evaluation, we fine-tune the full
encoder using the standard 1\% and 10\% labeled splits following prior works \cite{MAMP, GFP, SJEPA}.
For retrieval, we use $\ell_2$-normalized features and cosine similarity with
$k=1$.

\noindent\textbf{Scope of reported variability.}
Unless explicitly stated otherwise, the main results use one pre-training run.
The uncertainty in Table~\ref{tab:target_control} is the sample standard
deviation from three independently trained linear probes on one fixed
pre-trained checkpoint. The uncertainty in
Tables~\ref{tab:test_corruption}--\ref{tab:degraded_pretraining} is the sample
standard deviation over three independently sampled test-time corruptions.
Neither quantity estimates pre-training-run variability.

\section{Hierarchical Temporal Sampling}
\label{sec:sup_B}

SLiM uses hierarchical temporal sampling to construct semantically consistent
global and local views at different temporal granularities. For each view, we
first sample a temporal interval from the raw skeleton sequence and then
uniformly sample frames within that interval. If the sampled physical duration
is shorter than the target number of frames, we use linear temporal interpolation
to resize the clip to the desired resolution.

\noindent\textbf{Global view sampling.}
We generate two global views, $\mathbf{X}_{\mathrm{G1}}$ and
$\mathbf{X}_{\mathrm{G2}}$, by sampling two temporal intervals from the raw
sequence and resizing each to $T=64$ frames. The interval duration is randomly
sampled to cover 50\% to 100\% of the input sequence length:
$0.5 < (t_{\mathrm{end}}-t_{\mathrm{start}})/T_{\mathrm{input}} < 1.0$.
This allows each global view to capture the main action context while providing
view diversity for contrastive learning.

\noindent\textbf{Local view sampling.}
As illustrated in Fig.~\ref{fig:framework}, local views
$\mathbf{X}_{\mathrm{L1}}$, $\mathbf{X}_{\mathrm{L2}}$, and
$\mathbf{X}_{\mathrm{L3}}$ are sampled from the same temporal interval
$[t^1_{\mathrm{start}},t^1_{\mathrm{end}}]$ as the primary global anchor
$\mathbf{X}_{\mathrm{G1}}$. This anchoring reduces semantic misalignment between
global and local views. For each resolution, we sample two independent crops:
\begin{itemize}
    \item 32-frame views $\mathbf{X}_{\mathrm{L1}}$: crop duration ratio
    $p \in [0.35, 0.7]$.
    \item 16-frame views $\mathbf{X}_{\mathrm{L2}}$: crop duration ratio
    $p \in [0.15, 0.4]$.
    \item 8-frame views $\mathbf{X}_{\mathrm{L3}}$: crop duration ratio
    $p \in [0.05, 0.2]$.
\end{itemize}

\section{Semantic Tube Masking (STM)}
\label{sec:stm}

\noindent\textbf{Concept and motivation.}
Independent joint or patch masking can be solved by local interpolation because
neighboring skeleton joints are highly correlated in space and time. Semantic
Tube Masking (STM) reduces this shortcut by masking connected anatomical regions,
such as arms, legs, or torso, over consecutive frames. This forms
skeletal-temporal tubes and encourages the model to predict missing features
from broader body context and temporal motion.

\noindent\textbf{Anatomical joint grouping.}
For the standard 25-joint topology used by NTU-60~\cite{NTU60},
NTU-120~\cite{NTU120}, and PKU-MMD II~\cite{PKU}, we divide the body into five
semantic regions: trunk/head, left arm, right arm, left leg, and right leg.
Within a sampled region, STM selects a graph-connected joint span rather than
independent joints.

\noindent\textbf{Target ratio, tube proposals, and realized coverage.}
For each corrupted view, we sample a target ratio
$r\sim\mathcal{U}(0.5,0.9)$ and set the budget to
$B=\lfloor rN\rfloor$, where $N=N_TN_J$ and the default compact grid has
$N=8\times25=200$ tokens. We use $A_{\min}=8$ and $A_{\max}=0.5N$ for the
proposal area. Each proposal samples an anatomical group, a connected joint
span $S$ of width $w=|S|$, an area $A$ no larger than the remaining budget, and
a temporal extent
$h=\operatorname{clip}(\operatorname{round}(A/w),1,N_T)$. Its temporal start is
sampled uniformly. A proposal is accepted only when it adds at least one new
cell without exceeding $B$; generation stops after ten consecutive proposals
add no new cells. This acceptance rule means that the realized coverage
approaches the sampled target from below. Across 11,520 generated masks, the
realized ratio has mean 0.695, standard deviation 0.114, and range
$[0.485,0.895]$.

\noindent\textbf{Mask-token and positional-encoding order.}
Let $\widetilde{\mathbf{Z}}_p$ denote the linearly projected patch content
before the skeletal positional embedding is added. The implemented corrupted
tokens expand Eq.~\ref{eq:mask} as
\begin{equation}
    \mathbf{Z}_{p}^{\texttt{mask}}
    = (1-\mathcal{M})\odot\widetilde{\mathbf{Z}}_p
    + \mathcal{M}\odot\mathbf{e}_{\texttt{[mask]}}^{b}
    + \mathbf{E}_{\mathrm{skel}}^{b}.
    \label{eq:mask_order_supp}
\end{equation}
Thus, content replacement occurs before the skeletal positional embedding is
added, and masked positions retain their joint identities. Temporal locations
are subsequently supplied by RoPE inside self-attention.

\noindent\textbf{Interpretation of the main-paper STM ablation.}
In the first four rows of Table~\ref{tab:ablation_stm_saa}(a), $\times$ denotes
no masking, ``Limited'' denotes conventional independent joint masking, and
$\checkmark$ denotes STM. In the last three rows, both the global and local
views use STM; only the target-ratio policy changes among fixed $r=0.9$,
$r\sim\mathcal{U}(0,0.5)$, and the proposed
$r\sim\mathcal{U}(0.5,0.9)$. These intervals specify target ratios rather than
guaranteed realized coverage.

\noindent\textbf{Dual-role application.}
STM is used for both MFP and GLCL. For the primary global view, STM is always
applied to construct the corrupted input for masked feature prediction. For
local views in the GLCL branch, STM is applied stochastically with probability
$p=0.5$ as a structural perturbation.

\begin{algorithm}[t]
\caption{Semantic Tube Masking (STM)}\label{alg:stm}
\begin{algorithmic}[1]
\Statex \textbf{Input:} target ratio $r$, grid size $N_T\times N_J$,
semantic groups $\mathcal{G}$, proposal bounds $A_{\min},A_{\max}$
\State $B\gets\lfloor rN_TN_J\rfloor$;
$\mathcal{M}\gets\mathbf{0}_{N_T\times N_J}$; $C\gets0$; $F\gets0$
\While{$C<B$ \textbf{and} $F<10$}
    \State $P_{\max}\gets\min(B-C,A_{\max})$
    \If{$P_{\max}<A_{\min}$}
        \State \textbf{break}
    \EndIf
    \State Sample $G\in\mathcal{G}$ and a graph-connected span $S\subseteq G$;
    $w\gets|S|$
    \State Sample $A\sim\mathcal{U}(A_{\min},P_{\max})$
    \State $h\gets\operatorname{clip}(\operatorname{round}(A/w),1,N_T)$
    \State Sample $t\sim\mathcal{U}\{0,\ldots,N_T-h\}$
    \State $\widetilde{\mathcal{M}}\gets\mathcal{M}$;
    $\widetilde{\mathcal{M}}[t:t+h,S]\gets1$;
    $\widetilde C\gets\sum_{i,j}\widetilde{\mathcal{M}}[i,j]$
    \If{$C<\widetilde C\leq B$}
        \State $\mathcal{M}\gets\widetilde{\mathcal{M}}$; $C\gets\widetilde C$;
        $F\gets0$
    \Else
        \State $F\gets F+1$
    \EndIf
\EndWhile
\Statex \textbf{Output:} Semantic Tube Mask $\mathcal{M}$
\end{algorithmic}
\end{algorithm}

\section{Skeleton-Aware Augmentations (SAA)}
\label{sec:saa}

As illustrated in Fig.~\ref{fig:augmentation}, standard geometric augmentations
can distort the articulated structure of the human body and produce unreliable
contrastive pairs. Skeleton-Aware Augmentations (SAA) improve view diversity
while preserving basic anatomical consistency. Each augmentation is applied
independently with probability $p=0.5$.

\subsection{Skeleton-Aware Rotation}
\label{sec:sar}

Uniform rotation strategies often restrict all axes to a narrow range to avoid
unrealistic poses. We decouple vertical rotation from out-of-plane tilting. The
vertical angle $\beta$ is sampled from
$\mathcal{U}(-\theta_{\mathrm{vert}},\theta_{\mathrm{vert}})$ with
$\theta_{\mathrm{vert}}=180^\circ$, allowing full $360^\circ$ rotation around
the $Y$-axis. The tilt angles $\alpha$ and $\gamma$ along the non-gravity axes
$(X,Z)$ are sampled from
$\mathcal{U}(-\theta_{\mathrm{tilt}},\theta_{\mathrm{tilt}})$ with
$\theta_{\mathrm{tilt}}=30^\circ$ to avoid excessive out-of-plane tilting.

\subsection{Skeleton-Aware Mirroring}
\label{sec:sam}

Index swapping alone changes left-right semantics but does not perform geometric
reflection in coordinate space. We therefore combine coordinate reflection along
the lateral $X$-axis with left-right joint reassignment using a predefined
bilateral mapping $\mathcal{I}_{\mathrm{swap}}$. This keeps joint identity
consistent with the mirrored geometry.

\subsection{Bone-Aware Scaling}
\label{sec:bas}

Naive scaling of raw joint coordinates can distort local skeleton structure. We
instead operate in bone-vector space. Each bone is decomposed into a direction
and a length, and random scaling factors $s_G\sim\mathcal{U}(0.85,1.15)$ are
applied to bone lengths while preserving bone directions. Joint coordinates are
then reconstructed in topological order from the root joint trajectory.

\begin{algorithm}[t]
\caption{Skeleton-Aware Rotation}\label{alg:saa_rot}
\begin{algorithmic}[1]
\Statex \textbf{Input:} Skeleton sequence $\mathbf{X} \in \mathbb{R}^{C \times T \times J}$, tilt angle limit $\theta_{\text{tilt}}$, vertical rotation limit $\theta_{\text{vert}}$.
\Statex 
\Statex \textit{\textbf{Step 1.} Sample Rotation Angles:}
\State Sample $X$-axis tilt angle $\alpha \sim \mathcal{U}(-\theta_{\text{tilt}}, \theta_{\text{tilt}})$.
\State Sample $Y$-axis vertical angle $\beta \sim \mathcal{U}(-\theta_{\text{vert}}, \theta_{\text{vert}})$. \Comment{Typically large, e.g., $180^\circ$}
\State Sample $Z$-axis tilt angle $\gamma \sim \mathcal{U}(-\theta_{\text{tilt}}, \theta_{\text{tilt}})$.
\Statex 
\Statex \textit{\textbf{Step 2.} Construct Rotation Matrix:}
\State Compute 3D rotation matrices $\mathbf{R}_x(\alpha)$, $\mathbf{R}_y(\beta)$, and $\mathbf{R}_z(\gamma)$.
\State Combine into a single rotation matrix: $\mathbf{R} = \mathbf{R}_z \mathbf{R}_y \mathbf{R}_x \in \mathbb{R}^{3 \times 3}$.
\Statex 
\Statex \textit{\textbf{Step 3.} Apply Rotation:}
\State Transform the coordinates for all joints:
\Statex $\mathbf{X}_{\text{rot}}[:,t,j] = \mathbf{R} \mathbf{X}[:,t,j]$ for all $t \in \{0, \cdots, T-1\}$ and $j \in \{0, \cdots, J-1\}$.
\Statex 
\Statex \textbf{Output:} Rotated sequence $\mathbf{X}_{\text{rot}}$
\end{algorithmic}
\end{algorithm}

\begin{algorithm}[t]
\caption{Skeleton-Aware Mirroring}\label{alg:saa_mirror}
\begin{algorithmic}[1]
\Statex \textbf{Input:} Skeleton sequence $\mathbf{X} \in \mathbb{R}^{C \times T \times J}$, predefined left-right joint swapping mapping $\mathcal{I}_{\text{swap}}$.
\Statex 
\Statex \textit{\textbf{Step 1.} Joint Index Swapping:}
\State Swap the left and right anatomical joints based on the mapping: 
\Statex $\mathbf{X}_{\text{swap}}[:,:,j] = \mathbf{X}[:,:, \mathcal{I}_{\text{swap}}(j)]$ for all $j \in \{0, \cdots, J-1\}$.
\Statex 
\Statex \textit{\textbf{Step 2.} Spatial Reflection:}
\State Invert the coordinates along the $X$-axis (Horizontal mirroring):
\Statex $\mathbf{X}_{\text{mirror}}[0,:,:] = -\mathbf{X}_{\text{swap}}[0,:,:]$. \Comment{Assuming channel 0 corresponds to the $X$-axis}
\Statex 
\Statex \textbf{Output:} Mirrored sequence $\mathbf{X}_{\text{mirror}}$
\end{algorithmic}
\end{algorithm}

\begin{algorithm}[t]
\caption{Bone-Aware Scaling}\label{alg:saa_scale}
\begin{algorithmic}[1]
\Statex \textbf{Input:} Skeleton sequence $\mathbf{X} \in \mathbb{R}^{C \times T \times J}$, parent joint mapping $\mathcal{P}$, semantic joint groups $\mathcal{G}$, scale range $[s_{\text{min}}, s_{\text{max}}]$.
\Statex 
\Statex \textit{\textbf{Step 1.} Joint to Bone Conversion:}
\State The root bone is initialized to zero: $\mathbf{B}[:,:,0]=\mathbf{0}$.
\State Extract bone vectors $\mathbf{B}$ from joints $\mathbf{X}$. For each joint $j \in \{1, \cdots, J-1\}$:
\Statex $\mathbf{B}[:,:,j] = \mathbf{X}[:,:,j] - \mathbf{X}[:,:, \mathcal{P}(j)]$.
\Statex 
\Statex \textit{\textbf{Step 2.} Generate Independent Scale Factors:}
\State Initialize scale factor matrix $\mathbf{S} \in \mathbb{R}^{J}$.
\For{each group $G \in \mathcal{G}$}
    \State Sample a scale factor $s_{G} \sim \mathcal{U}(s_{\text{min}}, s_{\text{max}})$.
    \State Assign $s_{G}$ to all joints in group $G$: $\mathbf{S}[j] = s_{G}$ for $j \in G$.
\EndFor
\Statex 
\Statex \textit{\textbf{Step 3.} Apply Bone Scaling:}
\State Scale each bone vector:
\Statex $\mathbf{B}_{\text{scale}}[:, :, j] = \mathbf{B}[:, :, j] \odot \mathbf{S}[j]$. \Comment{$\odot$ denotes element-wise multiplication}
\Statex 
\Statex \textit{\textbf{Step 4.} Reconstruct Joints:}
\State Initialize $\mathbf{X}_{\text{scale}}$ and preserve the original root joint trajectory:
\Statex $\mathbf{X}_{\text{scale}}[:,:,0] = \mathbf{X}[:,:,0]$.
\For{each joint $j \in \{1, \cdots, J-1\}$ in topological order}
    \State Recover joint positions:
    \Statex \hspace{0.4cm} $\mathbf{X}_{\text{scale}}[:,:,j] = \mathbf{X}_{\text{scale}}[:,:, \mathcal{P}(j)] + \mathbf{B}_{\text{scale}}[:,:,j]$
\EndFor
\Statex 
\Statex \textbf{Output:} Scaled sequence $\mathbf{X}_{\text{scale}}$
\end{algorithmic}
\end{algorithm}

\section{Additional Results and Discussion}
\label{sec:sup_E}

\subsection{Token-Budget Study}

\begin{table}[t]
    \scriptsize
    \centering
    \caption{
    Token-budget configurations and average accuracy on NTU-60.
    \(T\) denotes the sampled clip length, \(P_T\) denotes the temporal patch size,
    and \(N_T=T/P_T\) is the number of temporal tokens. All settings use \(J=25\)
    joints and \(P_J=1\). Accuracy is averaged over the X-Sub and X-View protocols
    under linear evaluation. Inference GFLOPs are measured for the downstream
    encoder. SLiM-8 uses the default total batch size of 768 unless otherwise
    specified.
    }
    \label{tab:token_budget_avg}
    \resizebox{1.0\textwidth}{!}{%
    \begin{tabular}{lccccc}
        \toprule
        \textbf{Method}
        & \textbf{Clip} \(T\)
        & \textbf{Patch} \(P_T\)
        & \textbf{Tokens} \(N_T \times N_J\)
        & \textbf{Inference GFLOPs}
        & \textbf{Avg. Acc. (\%)} \\
        \midrule
        SkeletonMAE-8~\cite{SkeletonMAE} & 64  & 8 & \(8 \times 25\)  & 3.59  & 70.50 \\
        SkeletonMAE-15~\cite{SkeletonMAE} & 120 & 8 & \(15 \times 25\) & 8.36  & 74.00 \\
        SkeletonMAE-30~\cite{SkeletonMAE} & 120 & 4 & \(30 \times 25\) & 28.32 & 76.25 \\
        \midrule
        MAMP-8~\cite{MAMP} & 64  & 8 & \(8 \times 25\)  & 3.59  & 80.90 \\
        MAMP-15~\cite{MAMP} & 120 & 8 & \(15 \times 25\) & 8.36  & 85.50 \\
        MAMP-30~\cite{MAMP} & 120 & 4 & \(30 \times 25\) & 28.32 & 87.00 \\
        \midrule
        S-JEPA~\cite{SJEPA} & 120 & 4 & \(30 \times 25\) & 28.32 & 87.55 \\
        GFP~\cite{GFP} & 120 & 4 & \(30 \times 25\) & 28.32 & 88.95 \\
        \midrule
        \rowcolor{orange!12}
        \textbf{SLiM-8 (Batch 384)} & 64  & 8 & \(8 \times 25\)  & 3.59  & 89.60 \\
        \rowcolor{orange!12}
        \textbf{SLiM-8 (Batch 768)} & 64  & 8 & \(8 \times 25\)  & 3.59  & 90.55 \\
        \rowcolor{orange!12}
        \textbf{SLiM-15} & 120 & 8 & \(15 \times 25\) & 8.36  & 90.75 \\
        \rowcolor{orange!12}
        \textbf{SLiM-30} & 120 & 4 & \(30 \times 25\) & 28.32 & 90.90 \\
        \bottomrule
    \end{tabular}}
\end{table}

\noindent
Table~\ref{tab:token_budget_avg} provides the detailed configurations used for
the token-budget analysis in Fig.~\ref{fig:token}. For coordinate-reconstruction
MAE baselines, reducing the temporal token budget from \(30\) to \(8\) leads to
clear accuracy degradation. SkeletonMAE~\cite{SkeletonMAE} drops from 76.25\% to
70.50\%, and MAMP~\cite{MAMP} drops from 87.00\% to 80.90\%. This suggests that
coordinate- or trajectory-level reconstruction objectives rely strongly on dense
skeletal-temporal context. In contrast, SLiM remains stable across token budgets, achieving 90.55\%,
90.75\%, and 90.90\% with \(8\), \(15\), and \(30\) temporal tokens, respectively.
This shows that SLiM's complete design is compatible with compact
skeletal-temporal tokenization. The controlled analysis in
Sec.~\ref{sec:objective_controls} further shows that feature prediction alone
does not explain this difference; the form and stability of the target matter.
The additional SLiM-8 result with batch size 384 isolates the effect of training
batch size under the same \(8\times25\) token grid. Increasing the batch size to
768 improves the average accuracy from 89.60\% to 90.55\%, suggesting that
SLiM benefits from the larger SSL batch in this configuration. This comparison
isolates batch-size sensitivity, not aggregate pre-training cost.

\subsection{Detailed Cost Comparison}

\begin{table}[t]
    \scriptsize
    \centering
    \caption{
    Detailed cost comparison on NTU-60. \(N_T\times N_J\) denotes the encoder
    token grid after patchification. Inference GFLOPs are measured for the downstream
    encoder. Module-level GFLOPs are separated into student encoder, decoder, and
    target/auxiliary branches for one forward view. The target/auxiliary branch includes target generation
    networks for prior methods when used, and the EMA teacher forward for SLiM.
    These values do not aggregate SLiM's multi-view student/teacher forwards and
    therefore are not a total per-iteration pre-training comparison.
    Training resources are reported as implementation context and are not intended
    as a controlled hardware comparison.
    }
    \label{tab:flops_supp}
    \resizebox{\textwidth}{!}{%
    \begin{tabular}{lcccccccc}
        \toprule
        \multirow{2}{*}{\textbf{Method}} 
        & \textbf{Tokens}
        & \makecell{\textbf{Inference}\\\textbf{GFLOPs}}
        & \multicolumn{3}{c}{\textbf{Single-view module GFLOPs}}
        & \multicolumn{2}{c}{\textbf{Reported Training Resource}} \\
        \cmidrule(lr){2-2} \cmidrule(lr){3-3} \cmidrule(lr){4-6} \cmidrule(lr){7-8}
        & $N_T \times N_J$ 
        & Enc. 
        & Enc. & Dec. & Target/Aux. 
        & GPU Type & Batch Size \\
        \midrule
        SkeletonMAE~\cite{SkeletonMAE} 
        & $30 \times 25$ & 28.32 & 1.97 & 17.70 & -- 
        & 8$\times$A6000 & 64 \\
        MAMP~\cite{MAMP} 
        & $30 \times 25$ & 28.32 & 1.97 & 17.70 & -- 
        & 4$\times$RTX3090 & 128 \\
        S-JEPA~\cite{SJEPA} 
        & $30 \times 25$ & 28.32 & 1.97 & 17.70 & 28.32 
        & 8$\times$A100 & 256 \\
        GFP~\cite{GFP} 
        & $30 \times 25$ & 28.32 & 1.97 & 1.57 & 0.64 
        & -- & -- \\
        \rowcolor{orange!12}
        \textbf{SLiM (Ours)} 
        & $8 \times 25$ & \textbf{3.59} & 3.59 & -- & 3.59 
        & 4$\times$A6000 & 768 \\
        \bottomrule
    \end{tabular}}
\end{table}

\noindent
Table~\ref{tab:flops_supp} provides a module-level cost breakdown. Dense-token
MAE-based methods feed only visible tokens to the pre-training encoder, but use a
full-sequence decoder and, in some cases, an additional target module. SLiM
removes the coordinate-reconstruction decoder; its target/auxiliary entry is the
EMA-teacher forward on compact tokens. Because the methods use different view
counts and training pipelines, these single-view numbers should not be read as
aggregate pre-training-cost ratios. The controlled 7.89$\times$ claim concerns
the downstream encoder, where all methods process their full unmasked token grid.
In our implementation, SLiM is trained with a total batch size of 768 on
4 NVIDIA RTX A6000 GPUs.

\subsection{Transfer Learning Evaluation}

We further evaluate cross-dataset transferability on PKU-MMD II~\cite{PKU}.
The encoder is first pre-trained on the source dataset using self-supervised
learning, then frozen and evaluated on the target dataset with a linear
classifier. This setting measures whether the learned representation transfers
across datasets without end-to-end fine-tuning.

Table~\ref{tab:transfer} reports the results when pre-training on the X-Sub
splits of NTU-60~\cite{NTU60} and NTU-120~\cite{NTU120}, followed by linear
evaluation on PKU-MMD II X-Sub. SLiM achieves 72.7\% and 75.3\% accuracy when
pre-trained on NTU-60 and NTU-120, respectively, outperforming prior skeleton
SSL baselines under the same transfer setting. These results suggest that the
features learned by masked feature prediction, STM, and GLCL transfer effectively
across dataset distributions.

\begin{table}[t]
    \scriptsize
    \centering
    \caption{
    Transfer learning performance on PKU-MMD II. We pre-train on the X-Sub splits
    of NTU-60 or NTU-120 and evaluate frozen representations on PKU-MMD II X-Sub
    with a linear classifier. \textbf{Bold} and \underline{underline} indicate the best and second-best
    results, respectively.
    }
    \label{tab:transfer}
    \resizebox{0.4\textwidth}{!}{%
    \begin{tabular}{lcc}
        \toprule
        \multirow{2}{*}{\textbf{Method}} & \multicolumn{2}{c}{\textbf{To PKU-II}} \\ \cmidrule(lr){2-3} 
        & NTU-60 & NTU-120 \\
        \midrule
        LongT-GAN \cite{LongTGAN} & 44.8 & -- \\
        ISC \cite{ISC} & 51.1 & 52.3 \\
        CMD \cite{CMD} & 56.0 & 57.0 \\
        MacDiff \cite{MacDiff} & \underline{72.2} & 73.4 \\
        IGM \cite{IGM} & 59.8 & -- \\
        \midrule
        SkeletonMAE \cite{SkeletonMAE} & 58.4 & 61.0 \\
        MAMP \cite{MAMP} & 70.6 & 73.2 \\
        S-JEPA \cite{SJEPA} & 71.4 & \underline{74.2} \\
        \midrule
        \rowcolor{orange!15}
        \textbf{SLiM (Ours)} & \textbf{72.7} & \textbf{75.3} \\
        \bottomrule
    \end{tabular}}
\end{table}

\subsection{Controlled Analysis of Masked-Prediction Targets}
\label{sec:objective_controls}

The main paper compares the complementary MFP and GLCL losses. We additionally
isolate the target used within MFP. All variants in
Table~\ref{tab:target_control} use the same $8\times25$ transferred encoder,
GLCL branch, masking, optimization schedule, and 100-epoch pre-training budget;
only the MFP target, loss, and target-specific module change. Every
target-specific module is discarded before linear evaluation.

\begin{table}[t]
    \scriptsize
    \centering
    \caption{Matched comparison of masked-prediction targets on NTU-60. The
    uncertainty shown for the proposed variant is the sample standard deviation
    over three independently trained linear probes on one fixed pre-trained
    checkpoint; it is not pre-training-run variability.}
    \label{tab:target_control}
    \resizebox{1.0\textwidth}{!}{%
    \begin{tabular}{lllccc}
        \toprule
        \textbf{MFP target} & \textbf{Loss} & \textbf{Target-specific module}
        & \textbf{X-Sub} & \textbf{X-View} & \textbf{Avg.} \\
        \midrule
        Raw joint coordinates & MSE & 5-block decoder
        & 84.30 & 89.62 & 86.96 \\
        Teacher-projected features & MSE & None
        & 84.53 & 89.32 & 86.92 \\
        Teacher-projected features & MSE & Predictor head
        & 85.07 & 90.12 & 87.59 \\
        \rowcolor{orange!12}
        Teacher prototype distribution & Cross-entropy & None
        & $87.66\!\pm\!0.11$ & $92.68\!\pm\!0.09$ & \textbf{90.17} \\
        \bottomrule
    \end{tabular}}
\end{table}

Coordinate reconstruction and direct feature regression perform almost
identically (86.96 vs.\ 86.92 average accuracy). Thus, removing the full-token
coordinate decoder or replacing coordinates with features is not, by itself,
the source of the gain. The direct feature-MSE branch also shows a degenerate
training trajectory: its MFP loss decreases from 0.313 to
$3.3\times10^{-5}$, while the effective rank of the masked-position projected
features decreases to 1.12. For a diagnostic feature matrix with singular
values $\{\sigma_j\}$, we compute the entropy-based effective rank as
\begin{equation}
    r_{\mathrm{eff}}
    = \exp\!\left(-\sum_j q_j\log q_j\right),
    \qquad q_j=\frac{\sigma_j}{\sum_k\sigma_k}.
    \label{eq:effective_rank}
\end{equation}
An effective rank close to one indicates that the branch outputs occupy an
almost one-dimensional subspace. This degeneration is specific to the MFP
branch; GLCL continues to train the shared encoder, whose features remain
non-degenerate. A predictor head stabilizes feature regression and improves it
by 0.54/0.80 points on X-Sub/X-View, but remains 2.59/2.56 points below the
prototype objective. We therefore attribute the result to the complete
asymmetric prototype-assignment design---EMA teacher, stop-gradient, and
clean/corrupted-view asymmetry---rather than to cross-entropy alone. Feature
regression is not intrinsically unstable, as the predictor control also shows.

\subsection{Robustness to Missing and Noisy Joints}
\label{sec:robustness}

We distinguish \emph{declared missingness}, where unavailable joints are routed
through the learned mask token, from \emph{unflagged corruption}, where filled
coordinates are presented as valid input. We use the submitted 150-epoch
encoder and its clean-trained linear classifier without corruption-specific
training, tuning, or checkpoint selection. Corruption is applied only at test
time. Selected joints are removed for every frame of a clip, preventing
recovery by temporal interpolation. Reported uncertainty is the sample standard
deviation over three independently sampled corruption seeds.

\begin{table}[t]
\centering
\scriptsize
\caption{Test-time robustness on NTU-60. (a) Declared missing joints and
localization noise, where $\sigma$ is normalized by each sequence's mean bone
length. (b) Unflagged missing joints on X-Sub. The same selected joints are
absent throughout a clip. Each $\pm$ value is the sample standard deviation
over three corruption draws, not pre-training variability.}
\label{tab:test_corruption}
\begin{minipage}[t]{0.49\textwidth}
    \centering
    {\small (a) Declared missingness and noise}\\[2pt]
    \resizebox{\textwidth}{!}{%
    \begin{tabular}{lccc}
        \toprule
        \textbf{Condition} & \textbf{Severity} & \textbf{X-Sub} & \textbf{X-View} \\
        \midrule
        Clean & -- & 87.9 & 93.2 \\
        \midrule
        Declared missing & 3/25 & $87.86\!\pm\!0.06$ & $92.94\!\pm\!0.03$ \\
        Declared missing & 5/25 & $87.69\!\pm\!0.12$ & $92.68\!\pm\!0.09$ \\
        Declared missing & 8/25 & $87.23\!\pm\!0.01$ & $92.14\!\pm\!0.01$ \\
        \midrule
        Gaussian noise & $\sigma=0.01$ & $87.78\!\pm\!0.04$ & $93.18\!\pm\!0.02$ \\
        Gaussian noise & $\sigma=0.03$ & $87.71\!\pm\!0.05$ & $93.01\!\pm\!0.02$ \\
        Gaussian noise & $\sigma=0.05$ & $87.62\!\pm\!0.02$ & $92.84\!\pm\!0.03$ \\
        \bottomrule
    \end{tabular}}
\end{minipage}
\hfill
\begin{minipage}[t]{0.49\textwidth}
    \centering
    {\small (b) Missing-joint representation (X-Sub)}\\[2pt]
    \resizebox{\textwidth}{!}{%
    \begin{tabular}{lccc}
        \toprule
        \multirow{2}{*}{\textbf{Representation}} & \multicolumn{3}{c}{\textbf{Missing joints / 25}} \\
        \cmidrule(lr){2-4}
        & \textbf{3} & \textbf{5} & \textbf{8} \\
        \midrule
        Declared mask token & 87.86 & 87.69 & 87.23 \\
        Parent-joint fill & $81.30\!\pm\!0.29$ & $76.41\!\pm\!0.06$ & $66.51\!\pm\!0.07$ \\
        Zero fill & $72.20\!\pm\!0.18$ & $62.23\!\pm\!0.21$ & $49.63\!\pm\!0.23$ \\
        \bottomrule
    \end{tabular}}
\end{minipage}
\end{table}

With eight declared missing joints, accuracy remains 87.23/92.14, approximately
0.6/1.1 points below the unrounded clean scores. Localization noise up to 5\%
of mean bone length also causes a modest decrease. In contrast, treating filled
coordinates as valid leads to substantially larger degradation. X-View follows
the same trend: parent-joint fill decreases from 86.02 to 71.01, and zero fill
from 76.95 to 53.14, as the number of missing joints increases from three to
eight. SLiM therefore provides a mechanism for \emph{declared} missingness, but
is not inherently robust to arbitrary incorrect coordinates. A deployment would
need an upstream confidence or visibility signal that can be converted into the
corresponding token mask; this conversion has not been evaluated end-to-end with
a particular RGB pose estimator.

\subsection{Corruption-Specific Degraded Pre-training}
\label{sec:degraded_pretrain}

We further test whether joint degradation in the pre-training data can improve
robustness. For each pre-training clip, five of the 24 non-reference joints are
sampled once, zero-filled for all frames, and left unflagged. The reference joint
used by standard NTU preprocessing is never removed. We additionally add
Gaussian noise with $\sigma=0.03$ times the sequence-level mean bone length.
All other settings match the clean control. Because the degraded run reached 50
epochs during the response period, Table~\ref{tab:degraded_pretraining} compares
matched 50-epoch checkpoints. Each encoder uses its own linear classifier
trained and selected on clean data. The table reports NTU-60 X-Sub; each
$\pm$ value is the sample standard deviation over three test-time corruption
draws, and there is one pre-training run per condition.

\begin{table}[t]
    \scriptsize
    \centering
    \caption{Matched 50-epoch clean and degraded pre-training. Degraded
    pre-training uses persistent unflagged zero-fill of five non-reference joints
    plus Gaussian noise with $\sigma=0.03$ times mean bone length.}
    \label{tab:degraded_pretraining}
    \resizebox{0.82\textwidth}{!}{%
    \begin{tabular}{lccc}
        \toprule
        \textbf{Test condition} & \textbf{Clean pre-training}
        & \textbf{Degraded pre-training} & $\boldsymbol{\Delta}$ \\
        \midrule
        Clean test & 86.12 & 83.86 & $-2.26$ \\
        Declared missing, 3/25 & $85.65\!\pm\!0.07$ & $82.48\!\pm\!0.06$ & $-3.17$ \\
        Declared missing, 8/25 & $85.03\!\pm\!0.12$ & $80.95\!\pm\!0.12$ & $-4.08$ \\
        Parent-joint fill, 3/25 & $80.40\!\pm\!0.18$ & $77.73\!\pm\!0.10$ & $-2.67$ \\
        Parent-joint fill, 8/25 & $67.39\!\pm\!0.37$ & $67.01\!\pm\!0.37$ & $-0.38$ \\
        Zero fill, 3/25 & $74.18\!\pm\!0.12$ & $79.75\!\pm\!0.08$ & $+5.57$ \\
        Zero fill, 5/25 & $66.14\!\pm\!0.28$ & $78.21\!\pm\!0.35$ & $+12.07$ \\
        Zero fill, 8/25 & $54.73\!\pm\!0.06$ & $75.92\!\pm\!0.11$ & $+21.18$ \\
        Gaussian noise, $0.05\times$ bone length & $85.83\!\pm\!0.03$
        & $83.72\!\pm\!0.07$ & $-2.10$ \\
        \bottomrule
    \end{tabular}}
\end{table}

Degraded pre-training substantially improves the matched unflagged zero-fill
condition: with eight missing joints, accuracy increases from 54.73 to 75.92,
and the drop relative to each model's own clean score decreases from 31.39 to
7.94 points. The improvement is corruption-specific. Clean accuracy decreases
by 2.26 points on X-Sub and 2.24 points on X-View (91.22 to 88.98); declared
missingness becomes worse, and the benefit does not transfer to parent-joint
fill or stronger localization noise. We therefore present degraded pre-training
as a robustness analysis, not as a default modification to SLiM. When
missingness can be declared through confidence or visibility information,
routing those positions through the learned mask token remains preferable.

\section{Limitations}
\label{sec:limitations}

SLiM targets efficient skeleton representation learning under a compact
skeletal-temporal token budget. While this provides a favorable
accuracy--efficiency trade-off on standard benchmarks, actions requiring subtle
high-frequency motion or precise temporal ordering may benefit from denser
temporal tokens. Our empirical scope is 3D joint sequences from three benchmarks
with a fixed 25-joint topology. Adapting the manually defined STM groups to other
topologies and evaluating motion capture, articulated meshes, body-worn sensors,
or 2D poses from in-the-wild RGB videos remain future work; cross-modal
generality is not established here.

The robustness analysis further distinguishes known missingness from arbitrary
incorrect coordinates. Routing unavailable joints through the learned mask token
works well, but requires a reliable upstream confidence or visibility signal.
We have not evaluated confidence-to-mask conversion with a particular pose
estimator, and unflagged parent- or zero-filled joints can cause severe
degradation. Reproducing one unflagged corruption during pre-training improves
that matched condition, but costs approximately 2.25 clean-accuracy points and
does not generalize uniformly to other corruptions. Finally, the main results
come from one pre-training run. The reported 7.89$\times$ reduction concerns
downstream encoder GFLOPs, not aggregate multi-view pre-training cost or measured
latency; wall-clock behavior depends on hardware and implementation.

\section{Broader Impact}
\label{sec:broader}

Efficient skeleton representation learning may benefit applications such as
human-computer interaction, robotics, rehabilitation support, and assistive
systems by reducing the computational cost of action understanding. Since
skeletons are more compact than RGB videos, they may also reduce storage and
computation in some privacy-sensitive settings. At the same time, action recognition technologies can raise privacy and
surveillance concerns if deployed to monitor individuals without appropriate
consent, transparency, or governance. Although this work uses public benchmark
datasets and does not deploy a real-world system, practical deployment should
follow privacy-preserving data collection practices, obtain appropriate consent,
and evaluate potential biases caused by pose estimation quality, missing joints,
camera viewpoints, or domain-specific sensor conditions.

\section{Compute Resources and Existing Assets}
\label{sec:assets}

\noindent\textbf{Compute resources.}
SLiM is pre-trained on 4 NVIDIA RTX A6000 GPUs with a total batch size of 768.
The final model is pre-trained for 150 epochs, while the main ablation models
are pre-trained for 100 epochs; the exploratory degraded-pre-training comparison
uses matched 50-epoch checkpoints. We report single-view module GFLOPs and
downstream encoder inference GFLOPs, but do not interpret the former as an
aggregate multi-view pre-training comparison. Wall-clock training and inference
time may vary depending on hardware, implementation, batch size, data loading,
and mixed-precision settings.

\noindent\textbf{Existing datasets.}
We use NTU RGB+D 60~\cite{NTU60}, NTU RGB+D 120~\cite{NTU120}, and
PKU-MMD II~\cite{PKU} as public benchmark datasets for skeleton-based action
recognition. We cite the original dataset papers and follow their official
evaluation protocols. The datasets are not redistributed in this work;
users should obtain them from the official sources and follow the corresponding
terms of use specified by the dataset providers.


\clearpage
\bibliographystyle{plain}
\bibliography{neurips_2026}

\end{document}